\documentclass[aip,jap,reprint]{revtex4-2}

\draft 

\usepackage{amssymb,amsmath}
\usepackage{graphicx}
\usepackage{dcolumn}
\usepackage{bm}
\usepackage{color}
\usepackage{float}
\usepackage{hyperref}
\usepackage[version=4]{mhchem}
\usepackage{relsize}
\usepackage{comment}
\usepackage{media9}
\usepackage{algorithm}
\usepackage{algpseudocode}
\usepackage{soul}

\newcommand{\Ivan}{\color{black}}

\begin{document}

\title{Quantum-tunnelling deep neural network for {\Ivan optical illusion recognition}} 
\affiliation{Artificial Intelligence and Cyber Futures Institute, Charles Sturt University, Bathurst, NSW 2795, Australia\looseness=-1}
\author{Ivan S.~Maksymov}
\email[Ivan S.~Maksymov]{imaksymov@csu.edu.au}

\date{\today}

\begin{abstract}
The discovery of the quantum tunnelling (QT) effect---the transmission of particles through a high potential barrier---was one of the most impressive achievements of quantum mechanics made in the 1920s. Responding to the contemporary challenges, I introduce a deep neural network (DNN) architecture that processes information using the effect of QT. I demonstrate the ability of QT-DNN to recognise optical illusions like a human. {\Ivan Tasking QT-DNN to simulate human perception of the Necker cube and Rubin's vase, I provide arguments in favour of the superiority of QT-based activation functions over the activation functions optimised for modern applications in machine vision, also showing that, at the fundamental level, QT-DNN is closely related to biology-inspired DNNs and models based on the principles of quantum information processing.} 
\end{abstract}

\maketitle 

\section{Introduction}
\subsection{Can AI see optical illusions?}
How will a conscious robot \cite{Tak13, Sam16} see us and our world? Will it understand how humans behave in natural complex environments such as oceans and space \cite{Yam06, Cle17, Ni20} and make decisions in virtual reality systems \cite{Mni15, Kup23, Kha21} and video games \cite{Wan21}? Speaking broadly, will a robot differentiate lie from truth in media, social networks and politics \cite{Gal_book, Mak24_information, Gal24}? 

Biological vision is enabled by physical, physiological and psychological processes \cite{Feynman}. In turn, AI employs machine vision \cite{Kor20} and neural network models \cite{Kim17} to recognise and classify objects. Hence, even though machine vision shares certain features with biological vision \cite{Wen23}, in general AI and humans see two very different worlds \cite{Ngu15, Bak18, Pan21, Fea23, Che23}.

\begin{figure}[t]
 \includegraphics[width=8.5cm]{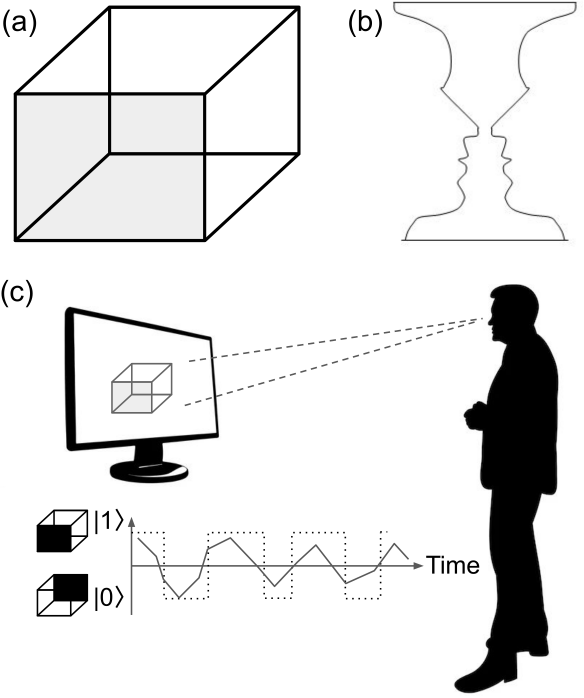}
 \caption{(a)~The Necker cube: The answer to the question `Is the shaded face of the cube at the front or at the rear?' will randomly switch between two stable perceptual states corresponding to the front ($|0\rangle$) and rear ($|1\rangle$) face of the cube. (b)~Rubin's vase: Do you see two people looking towards each other ($|0\rangle$) or a vase ($|1\rangle$)?' (c)~As per the traditional theory, the change from one perceptual state to another {\Ivan is binary (the dotted line), i.e.~it is from $|0\rangle$ to $|1\rangle$ and vice versa}. However, the current research works demonstrate that humans might see a superposition of the states $|0\rangle$ and $|1\rangle$ (the solid curve).\label{Fig_Necker}}
\end{figure}

For example, how would AI see the Necker cube \cite{Kor05, Bus12} and Rubin's vase \cite{Pin18, Kha21_1} (Fig.~\ref{Fig_Necker}a,~b)? While a human looking at these paradigmatic optical illusions \cite{Kor05} is able to report a random switching of their perception from one possible interpretation of the figure to another \cite{Bus12, Mak24_illusions} (Fig.~\ref{Fig_Necker}c), AI will struggle to reproduce the human perception in spite of the advances in simulation of human vision \cite{Ino94, Kud99, Mat18, Ara20, Sun21_1, Zha24, Zha24_1}.

\subsection{AI, psychology and quantum mechanics}
AI cannot see optical illusions like a human since its algorithms do not take into account psychological and neurological aspects of human vision \cite{Lon04, Kor05, Car14, Kha21_1}. {\Ivan An increasing number of research works have also suggested that this discrepancy is likely to arise due to the fundamental differences between artificial and biological neurons \cite{Bak18, Gom20, Kim21, Glo11, Bhu18, Jon21}. Although this problem has gained more attention in the recent years \cite{Wat18, Mel18, Kub21_1, Sha24}, the impact of visual and biological complexity of natural environments on the ability of AI to see like a human has not been fully appreciated \cite{Pax18, Agr20, Ni20}.} 

Experimental studies \cite{Gae98, Shi10, Pia17, Joo20, Mat23} demonstrate that the perception of the Necker cube and other ambiguous figures may not directly switch between the states $|0\rangle$ and $|1\rangle$ (the dotted line in Fig.~\ref{Fig_Necker}c) but can continuously oscillate between them (the solid curve in Fig.~\ref{Fig_Necker}c). Those results have been interpreted as the ability of humans to see a quantum-like superposition of $|0\rangle$ and $|1\rangle$ states \cite{Atm04, Khr06, Bus12, Pot22}. Indeed, in psychological experiments humans are asked to push an electric button every time their perception of the Necker cube switches between $|0\rangle$ and $|1\rangle$. While humans might see the cube in a superposition state \cite{Lon04, Mat23}, by pressing the button the observer `collapses' the superposition to one of the possible `classical' states \cite{Mak24_illusions}. This means that the ability of humans to access and process environmental information is limited, which may result in ambiguous interpretation. Subsequently, our perceptual system needs to disambiguate and interpret the available sensory information to construct stable and reliable percepts \cite{Lia18}. {\Ivan Yet, this suggests that AI can help humans enhance their perception and gain access to information that our senses and brain cannot process naturally. Such AI systems, including augmented cognition systems \cite{Sta09, Aga13} and decision-making support systems \cite{KPMG}, can be used to make high-quality administrative, political, financial and military decisions, taking in to account difficult to foresee scenarios of social and cultural interactions, geopolitical landscapes and battlefields in foreign lands \cite{Aga13}.} 

Further attempts to understand bistable perception using the methods of quantum mechanics \cite{Khr06, Bus12, Pot22} resulted in the suggestion to study the superposition of the perceptual states using a quantum oscillator model (QOM) \cite{Bus12}. Compared with Markov models \cite{Bus12}, QOM accounts for multiple outcomes while processing input data with a large set of constraints, thus describing human mental states \cite{Khr06, Pot22} more efficiently than the classical models.

The predictions of QOM are in good agreement with the results obtained using the models based on the quantum Zeno effect \cite{Atm04}. Besides, the ability of QOM to simulate the perception of optical illusions was improved \cite{Ben18, Mak24_illusions} by combining harmonic motion with the effect of quantum tunnelling (QT) through a potential barrier \cite{McQ97, Gri04}. Since both QOM and quantum Zeno effect underpin quantum neural network models \cite{Abb24}, the predictions made by QOM were additionally validated using a deep neural network (DNN) model \cite{Kim17, Mak24_illusions} combined with a quantum generator of truly random numbers \cite{Sym11}. 

\subsection{What is this article about?}
In this work, I suggest and theoretically validate a novel neuromorphic \cite{One22, Ye23} DNN model that processes information using the physical effect of QT. I showcase the ability of the QT-DNN model to enable AI to recognise practically important optical illusions, the perception of which should pave the way for the development of machine vision systems capable of recognising more complex visual effects used in fine arts and cinematography.

My findings can be used in AI systems and brain-computer interfaces aimed to enhance the performance of astronauts in long-duration spaceflight \cite{Cle17} and help airline pilots be aware of disorientation, loss of perspective and misinterpretation of data provided by flight instruments \cite{Deh22}. {\Ivan Since the analysis of ambiguous figures helps identify people who suffer from mild cognitive impairment, the algorithm proposed in this work can be employed in AI systems used by health professionals to diagnose debilitating diseases such as dementia \cite{Rue22}.} Interestingly enough, it can also help conduct research on gender since it has been demonstrated that fluidity of gender identity might be induced by optical illusions \cite{Tac20}. 
\begin{figure}[t]
 \includegraphics[width=8.5cm]{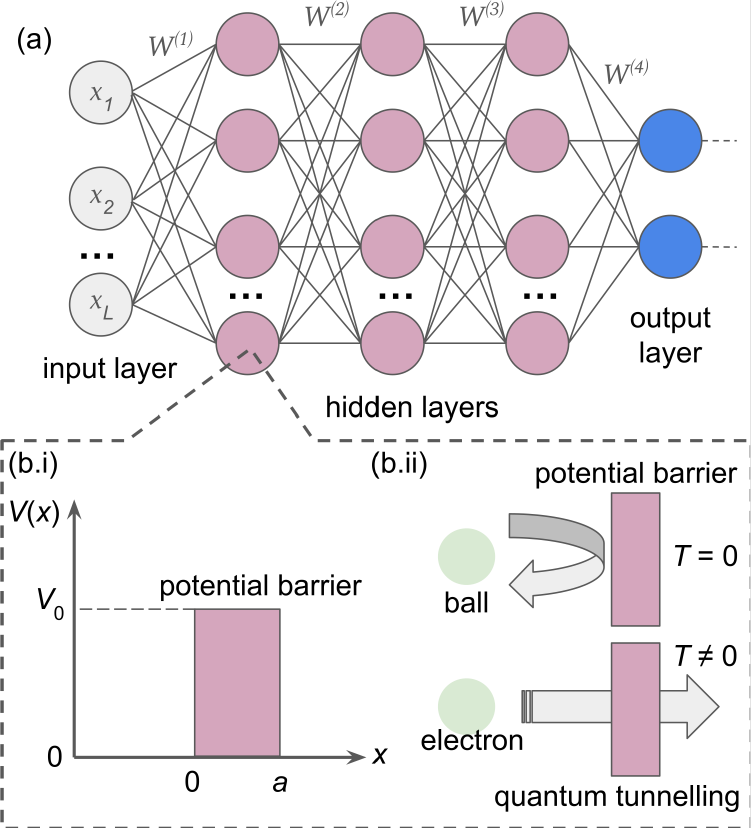}
 \caption{(a)~Sketch of QT-DNN structure. $W^{(n)}$ with $n=1\dots4$ are the matrices of the weights of the network connections. (b)~QT-DNN model employs the physical effect of QT as the activation function of its nodes. Panel~(b.i):~one-dimensional rectangular potential barrier of thickness $a$ and height $V_0$. Panel~(b.ii):~unlike in classical mechanics, in quantum mechanics there is a non-zero probability for an electron with energy $E<V_0$ to be transmitted through the barrier.\label{Fig1}}
\end{figure}

\section{QT-DNN architecture}

\subsection{QT-DNN algorithm}
Unless otherwise specified, the QT-DNN network used in this paper (Fig.~\ref{Fig1}a) consists of an input layer that has $L=100$ input nodes, three hidden layers each of which has $N=20$ nodes and an output layer that has $M=2$ output nodes that are used to classify the input dataset. The weights of the connections of the network are updated using a back-propagation training algorithm \cite{Kim17}. 

The activation function $\phi_{QT}$ of the nodes of the hidden layers is given {\Ivan by the algebraic expressions for the transmission coefficient $T$ of an electron penetrating a potential barrier (Fig.~\ref{Fig1}b). These expressions are well-known \cite{McQ97, Geo18} and they are presented in the Appendix section for the sake of self-consistency of the current paper.}

The output nodes of QT-DNN are governed by the Softmax function \cite{Kim17}
\begin{equation}
  \phi_{smax}(v_i) = \frac{\exp(v_i)}{\sum_{k=1}^{M} \exp(v_k)}\,,
  \label{eq:softmax}
\end{equation}
where $v_i$ is the weighted sum of input signals to the $i$th output node and $M$ is the number of the output nodes.

The network is trained and exploited as follows. First, I construct the output nodes that correspond to the correct answers to the training datasets. Then, I initialise the weights of the neural network in the range from --1 to 1 using a random number generator. Entering the input data $x_j$ and the corresponding training data points $d_i$, I calculate the error $e_i$ between the output $y_i$ and target $d_i$ as $e_i=d_i-y_i$. Then, propagating the output $\delta_i=e_i$ in the backward direction of the network, I compute the respective parameters $\delta_i^{(n)}$ of the hidden nodes using the equations $e_i^{(n)} = W^{{(n)}^\top}\delta_i$ and $\delta_i^{(n)} =\phi_{QT}^\prime\left(v_i^{(n)}\right)e_i^{(n)}$, where the index $n$ denotes the sequential number of the hidden layer, prime denotes the derivative of the activation function and $W^\top$ is the transpose of the matrix of weights corresponding to each relevant layer of the network. I continue the back-propagation process until the algorithm reaches the first hidden layer and then I update the weights using the learning rule $w_{ij}^{(n)}\gets w_{ij}^{(n)}+\Delta w_{ij}^{(n)}$, where $w_{ij}^{(n)}$ are the weights between an output node $i$ and input node $j$ of the $n$th layer and $\Delta w_{ij}^{(n)} =\alpha \delta_i^{(n)} x_j$. These computational steps are sequentially applied to all training data points.
\begin{figure}[t]
 \includegraphics[width=8.5cm]{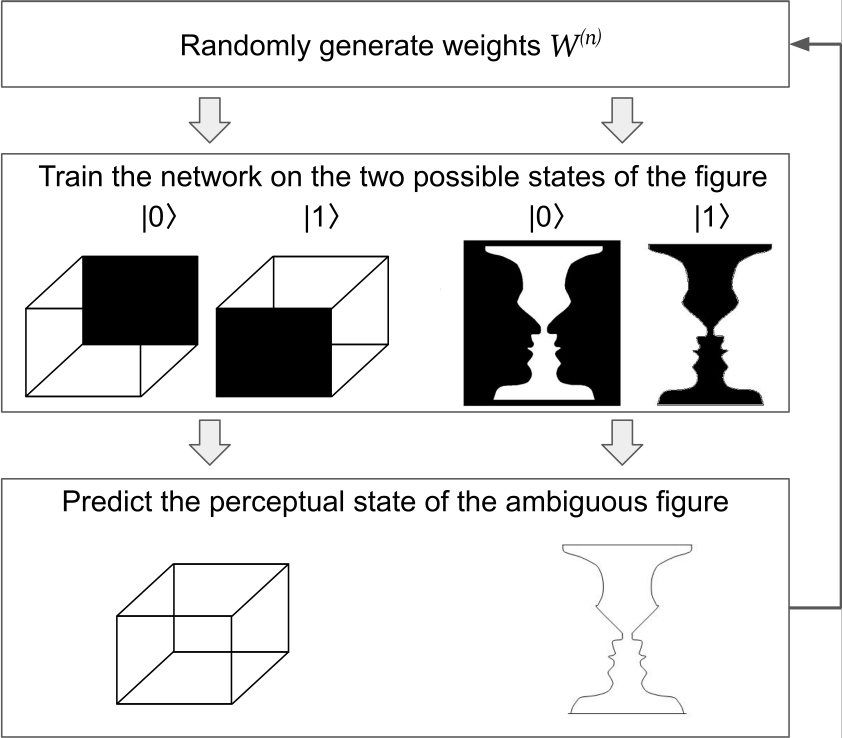}
 \caption{Neuromorphic algorithm involving the training of the network on distinguishable images of objects and its further exploitation aimed to recognise optical illusions.\label{Fig_test}}
\end{figure}

\subsection{Neuromorphic QT-DNN model of perception}
A biological brain is a nonlinear dynamical system that exhibits chaotic behaviour \cite{Kor03}. This property has motivated the development of neuromorphic AI that mimics the operation of the brain by exploiting nonlinear dynamical properties of diverse physical systems \cite{Tan19, Mar20, Mak23_review}.  

The principal components of the human vision system, including the retina and visual cortex, also exhibit nonlinear dynamical properties that can be used to create neuromorphic computers \cite{Wan20}. The visual input to the retina is modulated by eye blinks and movements, which effectively converts spatial information in the temporal one \cite{Yan24}, also playing an important role in cognition and visual perception \cite{Sak95, Ang20, Her24}. Yet, the dynamics of eye blink is also nonlinear and it may exhibit phase changes and chaotic behaviour \cite{Har18}, which are the processes that underpin the perception of optical illusions \cite{Gla15}. 

Based on these facts, it was demonstrated that the introduction of chaotic changes in the architecture of a neural network enables modelling the dynamics of information perception \cite{Ino94, Mak24_illusions}. Thus, I randomly generate the weights $W^{(n)}$, then I train QT-DNN on the $10\times10$\,pixel  unambiguous images of the Necker cube and then I exploit it to predict the perceptual state of the ambiguous Necker cube. This procedure is repeated in a loop to plot the dynamics of the perceived states (Figure~\ref{Fig_test}).

To create a more realistic model of quantum dynamics, I employ a physical generator of random numbers based on coherent quantum-optical processes \cite{Sym11} (the emission of coherent light can be associated with the effect of QT \cite{Par19}). Unlike a pseudo-random generator, a quantum generator produces truly random numbers \cite{Sym11}. This ensures that the software model of QT-DNN is not biased towards one of the possible perceptual states of optical illusions \cite{Ino94, Fan18}.

Rubin's vase training images had $20\times20$\,pixels. Although some works studying the Necker cube claim that similar results would be obtained for Rubin's vase \cite{Ben18}, recent research demonstrated that Rubin's vase has an increased contextual complexity \cite{Kha21_1}. Hence, I train QT-QNN using the figures with the shaded faces and vase, respectively, and then exploit it using a contour version of the drawing (Figure~\ref{Fig_test}). Both shaded and contour versions have been used in the literature \cite{Pin18} and they represent an intriguing benchmarking task, especially because the shaded training images are also ambiguous.

\section{Results}
Figure~\ref{Fig_Necker_result}a plots the simulated probability of perceiving $|0\rangle$ and $|1\rangle$ states of the Necker cube as a function of time (arbitrary time units are used in this work;~for a relevant discussion of the cognitive timescale see Refs.~\cite{Atm04, Kor12}). {\Ivan The nondimensionalised thickness of the potential barrier employed as the activation function of QT-DNN is $\sqrt{2mV_0}a/\hbar=$~0.5 (see the Appendix section for details; for a relevant discussion of the role of the  barrier parameters in the quantum models of cognition see Refs.~\cite{Mak24_illusions, Mak24_information}).} 

The probability curves were obtained as a result of 40~consecutive computational runs of the algorithm outlined in Fig.~\ref{Fig_test}. The states of the two output nodes of QT-DNN were recorded at the end of each run. Every pair of those data points was computed using a unique set of neural weights $W^{(n)}$. The same procedure was followed to simulate the perception of Rubin's vase (Fig.~\ref{Fig_Rubin_result}). The same values of $W^{(n)}$ were used in Fig.~\ref{Fig_Necker_result} and Fig.~\ref{Fig_Rubin_result}. 

In Fig.~\ref{Fig_Necker_result}a, we can observe a time-dependent switching between the fundamental perceptual states $|0\rangle$ and $|1\rangle$ of the Necker cube. {\Ivan In agreement with the predictions of the quantum cognition theory \cite{Bus12}, the switching between $|0\rangle$ and $|1\rangle$ is not binary (i.e.~it does not occur directly from $|0\rangle$ to $|1\rangle$ and vice versa) but it involves intermediate states that can be interpreted as a superposition of $|0\rangle$ and $|1\rangle$. At some instances of time (e.g., $T=17$), the model predicts an important, from the point of view of the quantum cognition theory, result where the probability of simultaneously perceiving the two possible states of the cube oscillates around 0.5 \cite{Bus12, Mak24_illusions}.  

The results produced by QT-DNN can be used to simulate the classical perception of the Necker cube, i.e.~the perception that admits only two binary states of the cube \cite{Mak24_illusions}. To that end, it can be assumed that the classical perception {\tt Percept}~=~1 when the quantum probability $P_{|1\rangle}\ge0.5$ and {\tt Percept}~=~0 when $P_{|1\rangle}<0.5$.
\begin{figure}[t]
 \includegraphics[width=8.5cm]{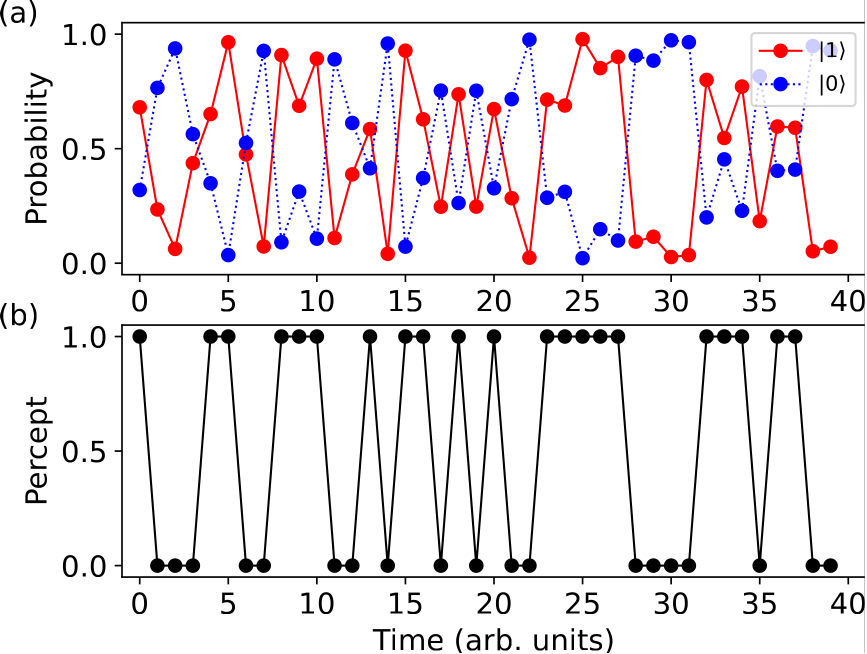}
 \caption{{\Ivan(a)~Perceptual switching data (with the lines serving as a guide to the eye) produced by QT-DNN trained on the Necker cube. The data points located in between 0 and 1 are in a superposition of the states $|0\rangle$ and $|1\rangle$. (b)~Classical perception computed using the procedure explained in the main text.}\label{Fig_Necker_result}}
\end{figure}

In Fig.~\ref{Fig_Necker_result}b, we can see that the classical perception curve consists of a series of square-like pulses of varying duration. This result should plausibly describe the human perception of the Necker cube. However, the classical result does not contain important information that can be used to better understand how humans control their perception of the illusion. In fact, the perception of the Necker cube is subjective and it may also be affected by the previous experience of the observer with ambiguous drawing and observer's attempts to force the switching to occur by focusing on different parts of the cube \cite{Lon04, Kor05} and voluntarily blinking \cite{Yan24}.       

As can be seen in Fig.~\ref{Fig_Rubin_result}, the pattern of the perception switching of Rubin's vase is similar to that of the Necker cube. However, QT-DNN predicts a higher switching frequency and longer periods of time corresponding to a superposition of the $|0\rangle$ and $|1\rangle$. This result is consistent with the theory that states that Rubin's vase combines an optical illusions with the perception of background \cite{Kha21_1}. Indeed, observers may interpret this figure not only as faces-vase but also as a white (black) vase on black (white) background, which gives them an additional point of reference that can be used to better disambiguate the drawing \cite{Wan17_1, Ngo08}.}
\begin{figure}[t]
 \includegraphics[width=8.5cm]{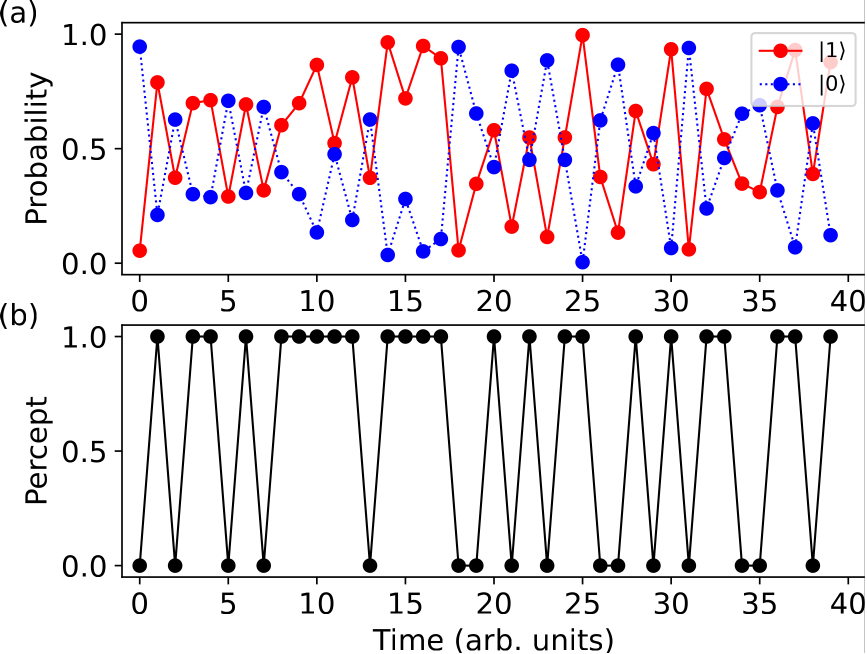}
 \caption{{\Ivan The same as in Fig.~\ref{Fig_Necker_result} but for Rubin's vase.}\label{Fig_Rubin_result}}
\end{figure}

\section{Discussion}
{\Ivan In this section, I demonstrate that, in the context of cognitive and decision-making DNN models, a QT-based activation function provides advantages over the standard activation functions employed in the field of machine learning \cite{Kim17}. I also show that, at the fundamental level, QT-DNN operates similarly to theoretical models of biological neural networks \cite{Day01, Glo11}. Although a solid link between the theory of quantum cognition and neuroscience is yet to be established, an extended discussion surveying the recent advances in the adjacent areas of quantum information and quantum consciousness (Sect.~\ref{Conclusion}) provides arguments in favour of its existence.
\begin{figure*}[t]
 \includegraphics[width=12.5cm]{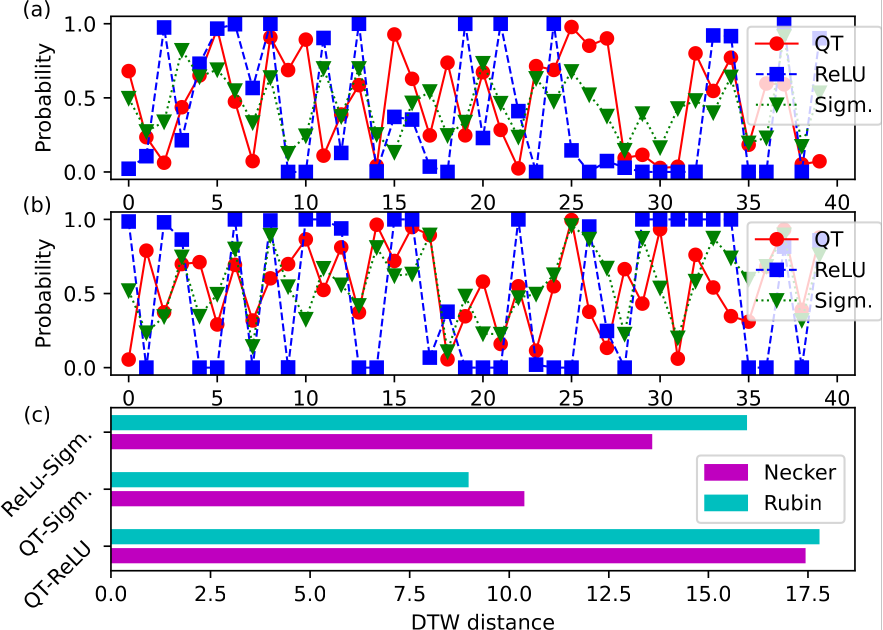}
 \caption{{\Ivan Probability of perceiving the $|1\rangle$ state of (a)~the Necker cube and (b)~Rubin's vase computed using DNN software that employs QT, ReLU and Sigmoid activation fucntions. (c)~Similarity between the perception switching curves calculated using the DTW algorithm.}\label{Fig_Necker_ActFunct}}
\end{figure*}

Several DNN models of human perception of optical illusions were proposed in the previous works \cite{Ino94, Kud99, Wat18, Mat18, Ara20, Sun21_1, Zha24, Zha24_1, Mak24_illusions}. Each model produced feasible results. However, no quantitative assessment of the accuracy of those models against psycho-physiological datasets has been carried out due to the unavailability of reliable human-generated data. Subsequently, as discussed above, experimental brain activity data have been used to evaluate the performance of the models. Thus, even though in its traditional formulation the quantum cognition theory does not imply any direct association with the function of a biological brain \cite{Bus12, Pot22}, it is conceivable that quantum cognition and neuroscience should be inter-related, at least in some aspects \cite{Sch05_1, Bus17, Nov24}.

Many DNN models adopted in the studies of optical illusions have been inspired by the advances made in the field of machine vision, thereby employing the advanced activation functions such as ReLU \cite{Wat18, Zha24_1, Mak24_illusions}. However, it has been recognised that the application of ReLU and other standard activation functions in DNN models of cognition and brain activity may lead to incorrect results \cite{Glo11, Bhu18, Jon21}. Yet, although Sigmoid-like activation function are also employed in DNN models \cite{Kim17}, their application in the studies of optical illusions appears to be motivated not by the knowledge from the field of machine learning but by the fact that similar functions have been used in certain models of biological neurons \cite{Day01, Glo11}.

To provide a unified comparison, in Fig.~\ref{Fig_Necker_ActFunct}a,~b I plot the probability of perceiving the $|1\rangle$ state of the Necker cube and Rubin's vase, respectively, computed by the DNN models that employ QT, ReLU and Sigmoid activation functions (all DNNs were trained under otherwise identical conditions). To quantify the similarity between the calculated switching curves, I employ a dynamic time warping (DTW) time-series analysis algorithm \cite{Sal07} that enables measuring similarity (distance) between temporal sequences that vary in speed \cite{Ols18}. The choice of DTW is justified because this algorithm has often been applied to human-generated time series such as speech audio signals and walking dynamics \cite{Ols18}. Thus, as shown in Fig.~\ref{Fig_Necker_ActFunct}c, the most similar results were produced by QT-DNN and Sigmoid-DNN but the largest distance was observed between the outputs of QT-DNN and ReLU-DNN.

The perception switching patterns produced by ReLU-DNN resembles the classical binary perception patterns shown in Fig.~\ref{Fig_Necker_result}c and Fig.~\ref{Fig_Rubin_result}c: the output of ReLU-DNN is predominantly a pure $|0\rangle$ or $|1\rangle$ state, with just a few data points corresponding to a superposition of $|0\rangle$ and $|1\rangle$. Thus, even though ReLU-DNN might have demonstrated an optimal result in terms of the performance metrics adopted in the field of machine learning, its output disagrees with the predictions a large and growing body of quantum models of perception. Moreover, in line with other relevant works \cite{Bak18, Gom20, Kim21}, this discrepancy may indicate that the use of ReLU in DNN models designed to recognise optical illusions is unjustified.    

On the other hand, a similarity of the outputs produced by QT-DNN and Sigmoid-DNN suggests that the effect of QT serves as a more adequate activation function for the analysis of psycho-physiological data. The following discussion provides additional arguments in support of this statement.

Firstly, as shown in Fig.~\ref{Fig3}, the shape of the QT activation function is similar to the shape of the neural activation function motivated by biological data \cite{Ben10, Glo11}. The functionality behind such a shape enables the model to satisfactorily account for the relation between the firing rate of a biological neuron and the input current, also considering the minimal time between two action potentials and other biological mechanisms \cite{Day01, Glo11}. 

Secondly, it has been demonstrated that neural networks based on a Sigmoid-like activation function can simulate such important operations of the quantum decision-making theory as conjunction fallacies (violation of the laws of the classical probability theory by human decision-making) \cite{Bus17}. While the authors of the cited paper might choose a Sigmoid-like activation function due its application in the models of biological neurons, their choice is also consistent with both regular techniques of probability estimation in social-economic research (e.g., with logistic regression \cite{Lip18}) as well as with novel theoretical approaches that analyse human behaviour through the prism of quantum physics \cite{Mak24_information1}.

Admittedly, the comparative analysis conducted above does not eliminate the need for a detailed benchmarking of the model outputs against rigorous experimental data. It is likely that such a comparison will be made in the future. Nevertheless, especially because further examples of the application of Sigmoid-like functions \cite{Zak00} and the effect of QT \cite{Ben18, Mak24_illusions} in quantum models of decision-making can be presented, it possible to claim that the results presented in this paper help establish a link between deep learning techniques and quantum cognition theories, also making us one step closer to the development of conscious AI, the potential benefits of which have been anticipated in the Introduction section.}

\section{Conclusions and Outlook\label{Conclusion}}
In this paper, I pointed out the possibility of employing the effect of quantum tunnelling as an activation function of artificial neural networks. I demonstrated that the so-constructed neural networks enable accurate modelling of human perception of optical illusions. 

During the peer-review process, my attention was brought to the fact that a digital computer running software that implements the QT-DNN algorithm technically remains a classical system, even though the model is based on the solution of a Schr{\"o}dinger equation and employs random number sets generated by a purely quantum physical system. The classical nature of QT-DNN software is inconsequential in the frameworks of sociophysics and quantum cognition theory, where, generally speaking, a model remains valid as long as the idealisations made in it and the assumptions about the data of interest hold true \cite{Mak24_information}. Indeed, it has been shown that empirical psychological data obtained in experiments aimed to reveal imprecision of human preference can be robustly explained using a classical electrodynamical model of an electron device based on the purely quantum phenomenon of spin-transfer torque \cite{Mak24_information1}.  

However, the distinction between classical and quantum systems becomes more prominent in the framework of the theories \cite{Geo18, Geo_book, Geo19, Geo20, Geo21, Geo22, Geo22_1, Geo22_2, Geo24} attempting to explain human consciousness and brain function from the point of view of the quantum information theory \cite{Chi19}. Although the quantum cognition theory and the quantum brain theory are formally independent one of another \cite{Bus12, Pot22}, they should converge at a certain point, resulting in a unified theory \cite{Sch05_1, Bus17, Khr20}. While the task of such a unification would be too ambitious for this paper, also going well beyond its scope, I note that a potential implementation of QT-DNN in hardware should result in a purely quantum neuromorphic computing system based on an analogue platform. Although a quantum neuromorphic system is different from a conventional quantum computing architecture \cite{Mar20_2}, at the conceptual level its hardware implementation should comply with the definition of `genuine quantum hardware' \cite{Chi19} identified in the framework of the quantum brain theory as an essential criterion for understanding consciousness \cite{Geo_book}.

\section*{Data Availability Statement}
The computational code that implements QT-DNN and relevant data can be found at \url{https://github.com/IvanMaksymov/DeepTunnel} ({accessed on 26 July 2024})


\appendix

\section{Quantum tunnelling as an activation function}
{\Ivan The electron tunnelling behaviour can be quantified by finding the transmission coefficient from the solution of
a time-independent Schr{\"o}dinger equation
\begin{equation}
  \label{eq:SE}
  \left[-\frac{\hbar^2}{2m}\frac{d^2}{d x^2} + V(x)\right]\psi(x) = E\psi(x)\,, 
\end{equation}
where $\psi(x)$ is a wave function, $m$ is the mass of the electron, $\hbar$ is Plank's constant and $E$ is the energy of the electron. Omitting the intermediate derivations \cite{McQ97}, I present the expressions for the probability of the electron transmission through the potential barrier with the profile $V(x)$ depicted in Fig.~\ref{Fig1}b.i.}

In classical mechanics, a counterpart of this physical system is a marble ball (Fig.~\ref{Fig1}b.ii). While a ball with energy $E<V_0$ cannot penetrate the barrier, an electron, behaving as a matter wave, has a non-zero probability of penetrating the barrier and continuing its motion on the other side. Similarly, for $E>V_0$, the electron may be reflected from the barrier with a non-zero probability.
\begin{figure}[t]
 \includegraphics[width=8.5cm]{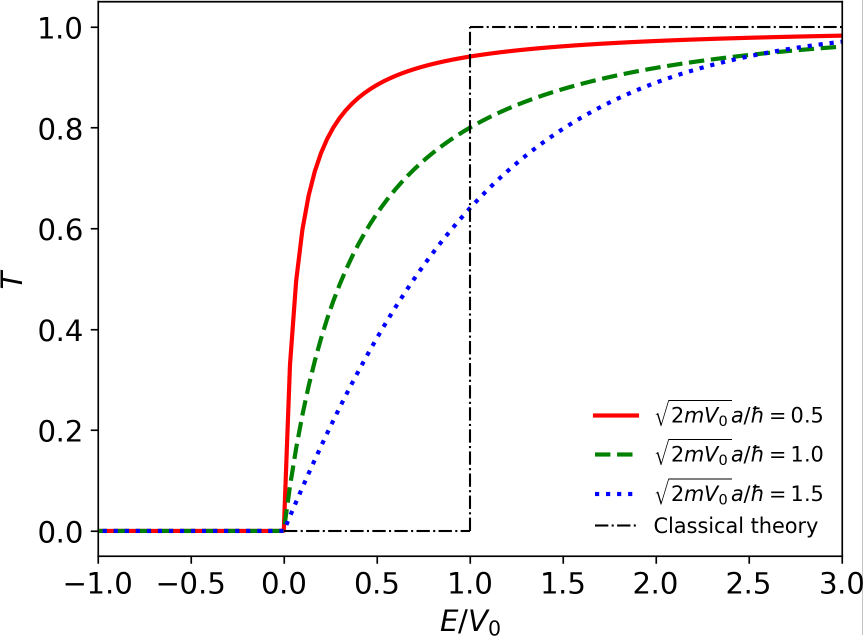}
 \caption{Transmission probability of an electron with energy~$E$ through a rectangular potential barrier with a nondimensionalised thickness $\sqrt{2mV_0}a/\hbar$. The straight dash-dotted lines denote the result predicted by the classical theory.\label{Fig3}}
\end{figure}

For electron energies smaller than the barrier height ($E<V_0$), there is a non-zero transmission probability   
\begin{equation}
  \label{eq:eq1}
  T\vert_{E<V_0} = \left(1-\beta\sinh^2(\kappa_1 a)\right)^{-1}\,, 
\end{equation}
where $\alpha=E-V_0$, $\beta=\dfrac{V_0^2}{4E\alpha}$ and $\kappa_1=\sqrt{-2m\alpha/\hbar^2}$. For $E>V_0$
\begin{equation}
  \label{eq:eq2}
  T\vert_{E>V_0} = \left(1+\beta\sin^2(\kappa a)\right)^{-1}\,, 
\end{equation}
where $\kappa=\sqrt{2m\alpha/\hbar^2}$. Finally, the expression for $E=V_0$ is obtained by taking the limit of $T$ as $E$ approaches $V_0$, resulting in
\begin{equation}
  \label{eq:eq3}
  T\vert_{E=V_0} = \left(1+\frac{ma^2V_0}{2\hbar^2}\right)^{-1}\,. 
\end{equation}

Since the derivatives of Eqs.~(\ref{eq:eq1}--\ref{eq:eq2}) with respect to $E$ are required for the DNN algorithm, I obtain
\begin{equation}
\begin{gathered}\label{eq:diff1}
  T^\prime\vert_{E<V_0}=-\beta\biggl(\frac{\sinh^2(\delta_1)}{E}+ \\\frac{\sinh^2(\delta_1)-\delta_1\cosh(\delta_1)\sinh(\delta_1)}{\alpha}\biggr)T^2\vert_{E<V_0}\,, \\
  T^\prime\vert_{E>V_0}=\beta\biggl(\frac{\sin^2(\delta)}{E}+\frac{\sin^2(\delta)-\delta\cos(\delta)\sin(\delta)}{\alpha}\biggr)T^2\vert_{E>V_0}\,, \\
  T^\prime\vert_{E=V_0}=\frac{4V_0a^4m^2+6a^2\hbar^2m}{3V_0^2a^4m^2+12V_0a^2\hbar^2m+12\hbar^4}\,,
\end{gathered}
\end{equation}
where $\delta=\kappa a$ and $\delta_1=\kappa_1a$.

For illustration, Fig.~\ref{Fig3} shows the probability of transmission of a single electron with energy $E$ through a potential barrier with the nondimensionalised thickness $\sqrt{2mV_0}a/\hbar$.

\bibliography{biblio}

\begin{thebibliography}{106}%
\makeatletter
\providecommand \@ifxundefined [1]{%
 \@ifx{#1\undefined}
}%
\providecommand \@ifnum [1]{%
 \ifnum #1\expandafter \@firstoftwo
 \else \expandafter \@secondoftwo
 \fi
}%
\providecommand \@ifx [1]{%
 \ifx #1\expandafter \@firstoftwo
 \else \expandafter \@secondoftwo
 \fi
}%
\providecommand \natexlab [1]{#1}%
\providecommand \enquote  [1]{``#1''}%
\providecommand \bibnamefont  [1]{#1}%
\providecommand \bibfnamefont [1]{#1}%
\providecommand \citenamefont [1]{#1}%
\providecommand \href@noop [0]{\@secondoftwo}%
\providecommand \href [0]{\begingroup \@sanitize@url \@href}%
\providecommand \@href[1]{\@@startlink{#1}\@@href}%
\providecommand \@@href[1]{\endgroup#1\@@endlink}%
\providecommand \@sanitize@url [0]{\catcode `\\12\catcode `\$12\catcode `\&12\catcode `\#12\catcode `\^12\catcode `\_12\catcode `\%12\relax}%
\providecommand \@@startlink[1]{}%
\providecommand \@@endlink[0]{}%
\providecommand \url  [0]{\begingroup\@sanitize@url \@url }%
\providecommand \@url [1]{\endgroup\@href {#1}{\urlprefix }}%
\providecommand \urlprefix  [0]{URL }%
\providecommand \Eprint [0]{\href }%
\providecommand \doibase [0]{https://doi.org/}%
\providecommand \selectlanguage [0]{\@gobble}%
\providecommand \bibinfo  [0]{\@secondoftwo}%
\providecommand \bibfield  [0]{\@secondoftwo}%
\providecommand \translation [1]{[#1]}%
\providecommand \BibitemOpen [0]{}%
\providecommand \bibitemStop [0]{}%
\providecommand \bibitemNoStop [0]{.\EOS\space}%
\providecommand \EOS [0]{\spacefactor3000\relax}%
\providecommand \BibitemShut  [1]{\csname bibitem#1\endcsname}%
\let\auto@bib@innerbib\@empty
\bibitem [{\citenamefont {Takeno}(2013)}]{Tak13}%
  \BibitemOpen
  \bibfield  {author} {\bibinfo {author} {\bibfnamefont {J.}~\bibnamefont {Takeno}},\ }\href@noop {} {\emph {\bibinfo {title} {Creation of a Conscious Robot: Mirror Image Cognition and Self-Awareness}}}\ (\bibinfo  {publisher} {CRC Press, Boca Raton},\ \bibinfo {year} {2013})\BibitemShut {NoStop}%
\bibitem [{\citenamefont {Samani}(2016)}]{Sam16}%
  \BibitemOpen
  \bibfield  {author} {\bibinfo {author} {\bibfnamefont {H.}~\bibnamefont {Samani}},\ }\href@noop {} {\emph {\bibinfo {title} {Cognitive Robotics}}}\ (\bibinfo  {publisher} {CRC Press, Boca Raton},\ \bibinfo {year} {2016})\BibitemShut {NoStop}%
\bibitem [{\citenamefont {Yamamoto}\ and\ \citenamefont {Yamamoto}(2006)}]{Yam06}%
  \BibitemOpen
  \bibfield  {author} {\bibinfo {author} {\bibfnamefont {S.}~\bibnamefont {Yamamoto}}\ and\ \bibinfo {author} {\bibfnamefont {M.}~\bibnamefont {Yamamoto}},\ }\bibfield  {title} {\enquote {\bibinfo {title} {Effects of the gravitational vertical on the visual perception of reversible figures},}\ }\href@noop {} {\bibfield  {journal} {\bibinfo  {journal} {Neurosci.~Res.}\ }\textbf {\bibinfo {volume} {55}},\ \bibinfo {pages} {218--121} (\bibinfo {year} {2006})}\BibitemShut {NoStop}%
\bibitem [{\citenamefont {Cl{\'e}ment}\ \emph {et~al.}(2015)\citenamefont {Cl{\'e}ment}, \citenamefont {Allaway}, \citenamefont {Demel}, \citenamefont {Golemis}, \citenamefont {Kindrat}, \citenamefont {Melinyshyn}, \citenamefont {Merali},\ and\ \citenamefont {Thirsk}}]{Cle17}%
  \BibitemOpen
  \bibfield  {author} {\bibinfo {author} {\bibfnamefont {G.}~\bibnamefont {Cl{\'e}ment}}, \bibinfo {author} {\bibfnamefont {H.~C.~M.}\ \bibnamefont {Allaway}}, \bibinfo {author} {\bibfnamefont {M.}~\bibnamefont {Demel}}, \bibinfo {author} {\bibfnamefont {A.}~\bibnamefont {Golemis}}, \bibinfo {author} {\bibfnamefont {A.~N.}\ \bibnamefont {Kindrat}}, \bibinfo {author} {\bibfnamefont {A.~N.}\ \bibnamefont {Melinyshyn}}, \bibinfo {author} {\bibfnamefont {T.}~\bibnamefont {Merali}},\ and\ \bibinfo {author} {\bibfnamefont {R.}~\bibnamefont {Thirsk}},\ }\bibfield  {title} {\enquote {\bibinfo {title} {Long-duration spaceflight increases depth ambiguity of reversible perspective figures},}\ }\href {https://doi.org/10.1371/journal.pone.0132317} {\bibfield  {journal} {\bibinfo  {journal} {PLOS ONE}\ }\textbf {\bibinfo {volume} {10}} (\bibinfo {year} {2015}),\ 10.1371/journal.pone.0132317}\BibitemShut {NoStop}%
\bibitem [{\citenamefont {Ni}\ \emph {et~al.}()\citenamefont {Ni}, \citenamefont {Du}, \citenamefont {Wang},\ and\ \citenamefont {Jiao}}]{Ni20}%
  \BibitemOpen
  \bibfield  {author} {\bibinfo {author} {\bibfnamefont {Y.}~\bibnamefont {Ni}}, \bibinfo {author} {\bibfnamefont {Z.}~\bibnamefont {Du}}, \bibinfo {author} {\bibfnamefont {S.}~\bibnamefont {Wang}},\ and\ \bibinfo {author} {\bibfnamefont {F.}~\bibnamefont {Jiao}},\ }\enquote {\bibinfo {title} {Rectification of driver’s visual illusion on continuous downhill of urban underwater tunnel},}\ in\ \href {https://doi.org/10.1061/9780784482933.318} {\emph {\bibinfo {booktitle} {CICTP 2020}}},\ pp.\ \bibinfo {pages} {3692--3704}\BibitemShut {NoStop}%
\bibitem [{\citenamefont {Mnih}\ \emph {et~al.}(2015)\citenamefont {Mnih}, \citenamefont {Kavukcuoglu}, \citenamefont {Silver}, \citenamefont {Rusu}, \citenamefont {Veness}, \citenamefont {Bellemare}, \citenamefont {Graves}, \citenamefont {Riedmiller}, \citenamefont {Fidjeland}, \citenamefont {Ostrovski}, \citenamefont {Petersen}, \citenamefont {Beattie}, \citenamefont {Sadik}, \citenamefont {Antonoglou}, \citenamefont {King}, \citenamefont {Kumaran}, \citenamefont {Wierstra}, \citenamefont {Legg},\ and\ \citenamefont {Hassabis}}]{Mni15}%
  \BibitemOpen
  \bibfield  {author} {\bibinfo {author} {\bibfnamefont {V.}~\bibnamefont {Mnih}}, \bibinfo {author} {\bibfnamefont {K.}~\bibnamefont {Kavukcuoglu}}, \bibinfo {author} {\bibfnamefont {D.}~\bibnamefont {Silver}}, \bibinfo {author} {\bibfnamefont {A.~A.}\ \bibnamefont {Rusu}}, \bibinfo {author} {\bibfnamefont {J.}~\bibnamefont {Veness}}, \bibinfo {author} {\bibfnamefont {M.~G.}\ \bibnamefont {Bellemare}}, \bibinfo {author} {\bibfnamefont {A.}~\bibnamefont {Graves}}, \bibinfo {author} {\bibfnamefont {M.}~\bibnamefont {Riedmiller}}, \bibinfo {author} {\bibfnamefont {A.~K.}\ \bibnamefont {Fidjeland}}, \bibinfo {author} {\bibfnamefont {G.}~\bibnamefont {Ostrovski}}, \bibinfo {author} {\bibfnamefont {S.}~\bibnamefont {Petersen}}, \bibinfo {author} {\bibfnamefont {C.}~\bibnamefont {Beattie}}, \bibinfo {author} {\bibfnamefont {A.}~\bibnamefont {Sadik}}, \bibinfo {author} {\bibfnamefont {I.}~\bibnamefont {Antonoglou}}, \bibinfo {author} {\bibfnamefont {H.}~\bibnamefont {King}}, \bibinfo {author} {\bibfnamefont
  {D.}~\bibnamefont {Kumaran}}, \bibinfo {author} {\bibfnamefont {D.}~\bibnamefont {Wierstra}}, \bibinfo {author} {\bibfnamefont {S.}~\bibnamefont {Legg}},\ and\ \bibinfo {author} {\bibfnamefont {D.}~\bibnamefont {Hassabis}},\ }\bibfield  {title} {\enquote {\bibinfo {title} {Human-level control through deep reinforcement learning},}\ }\href@noop {} {\bibfield  {journal} {\bibinfo  {journal} {Nature}\ }\textbf {\bibinfo {volume} {518}},\ \bibinfo {pages} {529--533} (\bibinfo {year} {2015})}\BibitemShut {NoStop}%
\bibitem [{\citenamefont {Kuperwajs}, \citenamefont {Sch{\"u}tt},\ and\ \citenamefont {Ma}(2023)}]{Kup23}%
  \BibitemOpen
  \bibfield  {author} {\bibinfo {author} {\bibfnamefont {I.}~\bibnamefont {Kuperwajs}}, \bibinfo {author} {\bibfnamefont {H.~H.}\ \bibnamefont {Sch{\"u}tt}},\ and\ \bibinfo {author} {\bibfnamefont {W.~J.}\ \bibnamefont {Ma}},\ }\bibfield  {title} {\enquote {\bibinfo {title} {Using deep neural networks as a guide for modeling human planning},}\ }\href@noop {} {\bibfield  {journal} {\bibinfo  {journal} {Sci.~Rep.}\ }\textbf {\bibinfo {volume} {13}},\ \bibinfo {pages} {20269} (\bibinfo {year} {2023})}\BibitemShut {NoStop}%
\bibitem [{\citenamefont {Khalid}\ and\ \citenamefont {Iida}(2021)}]{Kha21}%
  \BibitemOpen
  \bibfield  {author} {\bibinfo {author} {\bibfnamefont {M.~N.~A.}\ \bibnamefont {Khalid}}\ and\ \bibinfo {author} {\bibfnamefont {H.}~\bibnamefont {Iida}},\ }\bibfield  {title} {\enquote {\bibinfo {title} {Objectivity and subjectivity in games: Understanding engagement and addiction mechanism},}\ }\href@noop {} {\bibfield  {journal} {\bibinfo  {journal} {IEEE Access}\ }\textbf {\bibinfo {volume} {9}},\ \bibinfo {pages} {65187--65205} (\bibinfo {year} {2021})}\BibitemShut {NoStop}%
\bibitem [{\citenamefont {Wang}\ \emph {et~al.}(2021)\citenamefont {Wang}, \citenamefont {Xu}, \citenamefont {Wang}, \citenamefont {Huang}, \citenamefont {Chang}, \citenamefont {Cheng}, \citenamefont {Lin},\ and\ \citenamefont {Cheng}}]{Wan21}%
  \BibitemOpen
  \bibfield  {author} {\bibinfo {author} {\bibfnamefont {P.-Y.~C.}\ \bibnamefont {Wang}}, \bibinfo {author} {\bibfnamefont {C.-H.}\ \bibnamefont {Xu}}, \bibinfo {author} {\bibfnamefont {P.-Y.}\ \bibnamefont {Wang}}, \bibinfo {author} {\bibfnamefont {H.-Y.}\ \bibnamefont {Huang}}, \bibinfo {author} {\bibfnamefont {Y.-W.}\ \bibnamefont {Chang}}, \bibinfo {author} {\bibfnamefont {J.-H.}\ \bibnamefont {Cheng}}, \bibinfo {author} {\bibfnamefont {Y.-H.}\ \bibnamefont {Lin}},\ and\ \bibinfo {author} {\bibfnamefont {L.-P.}\ \bibnamefont {Cheng}},\ }\bibfield  {title} {\enquote {\bibinfo {title} {Game illusionization: A workflow for applying optical illusions to video games},}\ }in\ \href {https://doi.org/10.1145/3472749.3474824} {\emph {\bibinfo {booktitle} {The 34th Annual ACM Symposium on User Interface Software and Technology}}}\ (\bibinfo  {publisher} {Association for Computing Machinery},\ \bibinfo {address} {New York, NY, USA},\ \bibinfo {year} {2021})\ pp.\ \bibinfo {pages} {1326--1344}\BibitemShut {NoStop}%
\bibitem [{\citenamefont {Galam}(2012)}]{Gal_book}%
  \BibitemOpen
  \bibfield  {author} {\bibinfo {author} {\bibfnamefont {S.}~\bibnamefont {Galam}},\ }\href@noop {} {\emph {\bibinfo {title} {Sociophysics: A Physicist's Modeling of Psycho-political Phenomena}}}\ (\bibinfo  {publisher} {Springer New York},\ \bibinfo {year} {2012})\BibitemShut {NoStop}%
\bibitem [{\citenamefont {Maksymov}\ and\ \citenamefont {Pogrebna}(2024{\natexlab{a}})}]{Mak24_information}%
  \BibitemOpen
  \bibfield  {author} {\bibinfo {author} {\bibfnamefont {I.~S.}\ \bibnamefont {Maksymov}}\ and\ \bibinfo {author} {\bibfnamefont {G.}~\bibnamefont {Pogrebna}},\ }\bibfield  {title} {\enquote {\bibinfo {title} {Quantum-mechanical modelling of asymmetric opinion polarisation in social networks},}\ }\href@noop {} {\bibfield  {journal} {\bibinfo  {journal} {Information}\ }\textbf {\bibinfo {volume} {15}},\ \bibinfo {pages} {170} (\bibinfo {year} {2024}{\natexlab{a}})}\BibitemShut {NoStop}%
\bibitem [{\citenamefont {Galam}(2024)}]{Gal24}%
  \BibitemOpen
  \bibfield  {author} {\bibinfo {author} {\bibfnamefont {S.}~\bibnamefont {Galam}},\ }\bibfield  {title} {\enquote {\bibinfo {title} {Fake news: ``no ban, no spread—with sequestration''},}\ }\href@noop {} {\bibfield  {journal} {\bibinfo  {journal} {Physics}\ }\textbf {\bibinfo {volume} {6}},\ \bibinfo {pages} {859--876} (\bibinfo {year} {2024})}\BibitemShut {NoStop}%
\bibitem [{\citenamefont {Feynman}, \citenamefont {Leighton},\ and\ \citenamefont {Sands}(2010)}]{Feynman}%
  \BibitemOpen
  \bibfield  {author} {\bibinfo {author} {\bibfnamefont {R.~P.}\ \bibnamefont {Feynman}}, \bibinfo {author} {\bibfnamefont {R.~B.}\ \bibnamefont {Leighton}},\ and\ \bibinfo {author} {\bibfnamefont {M.}~\bibnamefont {Sands}},\ }\href@noop {} {\emph {\bibinfo {title} {{The Feynman lectures on physics. Mechanisms of Seeing}}}},\ Vol.~\bibinfo {volume} {3}\ (\bibinfo  {publisher} {Basic Books},\ \bibinfo {address} {New York, NY},\ \bibinfo {year} {2010})\ Chap.~\bibinfo {chapter} {36}\BibitemShut {NoStop}%
\bibitem [{\citenamefont {Kortli}\ \emph {et~al.}(2020)\citenamefont {Kortli}, \citenamefont {Jridi}, \citenamefont {Al~Falou},\ and\ \citenamefont {Atri}}]{Kor20}%
  \BibitemOpen
  \bibfield  {author} {\bibinfo {author} {\bibfnamefont {Y.}~\bibnamefont {Kortli}}, \bibinfo {author} {\bibfnamefont {M.}~\bibnamefont {Jridi}}, \bibinfo {author} {\bibfnamefont {A.}~\bibnamefont {Al~Falou}},\ and\ \bibinfo {author} {\bibfnamefont {M.}~\bibnamefont {Atri}},\ }\bibfield  {title} {\enquote {\bibinfo {title} {Face recognition systems: a survey},}\ }\href@noop {} {\bibfield  {journal} {\bibinfo  {journal} {Sensors}\ }\textbf {\bibinfo {volume} {20}},\ \bibinfo {pages} {342} (\bibinfo {year} {2020})}\BibitemShut {NoStop}%
\bibitem [{\citenamefont {Kim}(2017)}]{Kim17}%
  \BibitemOpen
  \bibfield  {author} {\bibinfo {author} {\bibfnamefont {P.}~\bibnamefont {Kim}},\ }\href@noop {} {\emph {\bibinfo {title} {MATLAB Deep Learning With Machine Learning, Neural Networks and Artificial Intelligence}}}\ (\bibinfo  {publisher} {Apress Berkeley, CA},\ \bibinfo {year} {2017})\BibitemShut {NoStop}%
\bibitem [{\citenamefont {Wencheng}\ \emph {et~al.}(2023)\citenamefont {Wencheng}, \citenamefont {Ge}, \citenamefont {Zuo}, \citenamefont {Chen}, \citenamefont {Qin},\ and\ \citenamefont {Zuxiang}}]{Wen23}%
  \BibitemOpen
  \bibfield  {author} {\bibinfo {author} {\bibfnamefont {W.}~\bibnamefont {Wencheng}}, \bibinfo {author} {\bibfnamefont {Y.}~\bibnamefont {Ge}}, \bibinfo {author} {\bibfnamefont {Z.}~\bibnamefont {Zuo}}, \bibinfo {author} {\bibfnamefont {L.}~\bibnamefont {Chen}}, \bibinfo {author} {\bibfnamefont {X.}~\bibnamefont {Qin}},\ and\ \bibinfo {author} {\bibfnamefont {L.}~\bibnamefont {Zuxiang}},\ }\bibfield  {title} {\enquote {\bibinfo {title} {Visual number sense for real-world scenes shared by deep neural networks and humans},}\ }\href@noop {} {\bibfield  {journal} {\bibinfo  {journal} {Heliyon}\ }\textbf {\bibinfo {volume} {9}},\ \bibinfo {pages} {e18517} (\bibinfo {year} {2023})}\BibitemShut {NoStop}%
\bibitem [{\citenamefont {Nguyen}, \citenamefont {Yosinski},\ and\ \citenamefont {Clune}(2015)}]{Ngu15}%
  \BibitemOpen
  \bibfield  {author} {\bibinfo {author} {\bibfnamefont {A.}~\bibnamefont {Nguyen}}, \bibinfo {author} {\bibfnamefont {J.}~\bibnamefont {Yosinski}},\ and\ \bibinfo {author} {\bibfnamefont {J.}~\bibnamefont {Clune}},\ }\bibfield  {title} {\enquote {\bibinfo {title} {Deep neural networks are easily fooled: High confidence predictions for unrecognizable images},}\ }in\ \href {https://doi.org/10.1109/CVPR.2015.7298640} {\emph {\bibinfo {booktitle} {2015 IEEE Conference on Computer Vision and Pattern Recognition (CVPR)}}}\ (\bibinfo {year} {2015})\ pp.\ \bibinfo {pages} {427--436}\BibitemShut {NoStop}%
\bibitem [{\citenamefont {Baker}\ \emph {et~al.}(2018)\citenamefont {Baker}, \citenamefont {Erlikhman}, \citenamefont {Kellman},\ and\ \citenamefont {Lu}}]{Bak18}%
  \BibitemOpen
  \bibfield  {author} {\bibinfo {author} {\bibfnamefont {N.}~\bibnamefont {Baker}}, \bibinfo {author} {\bibfnamefont {G.}~\bibnamefont {Erlikhman}}, \bibinfo {author} {\bibfnamefont {P.}~\bibnamefont {Kellman}},\ and\ \bibinfo {author} {\bibfnamefont {H.}~\bibnamefont {Lu}},\ }\bibfield  {title} {\enquote {\bibinfo {title} {Deep convolutional networks do not perceive illusory contours},}\ }\href@noop {} {\bibfield  {journal} {\bibinfo  {journal} {Proceedings of the Annual Meeting of the Cognitive Science Society}\ }\textbf {\bibinfo {volume} {40}},\ \bibinfo {pages} {1310--1315} (\bibinfo {year} {2018})}\BibitemShut {NoStop}%
\bibitem [{\citenamefont {Pang}\ \emph {et~al.}(2021)\citenamefont {Pang}, \citenamefont {O’May}, \citenamefont {Choksi},\ and\ \citenamefont {VanRullen}}]{Pan21}%
  \BibitemOpen
  \bibfield  {author} {\bibinfo {author} {\bibfnamefont {Z.}~\bibnamefont {Pang}}, \bibinfo {author} {\bibfnamefont {C.~B.}\ \bibnamefont {O’May}}, \bibinfo {author} {\bibfnamefont {B.}~\bibnamefont {Choksi}},\ and\ \bibinfo {author} {\bibfnamefont {R.}~\bibnamefont {VanRullen}},\ }\bibfield  {title} {\enquote {\bibinfo {title} {Predictive coding feedback results in perceived illusory contours in a recurrent neural network},}\ }\href@noop {} {\bibfield  {journal} {\bibinfo  {journal} {Neural Netw.}\ }\textbf {\bibinfo {volume} {144}},\ \bibinfo {pages} {164--175} (\bibinfo {year} {2021})}\BibitemShut {NoStop}%
\bibitem [{\citenamefont {Feather}\ \emph {et~al.}(2023)\citenamefont {Feather}, \citenamefont {Leclerc}, \citenamefont {Madry},\ and\ \citenamefont {McDermott}}]{Fea23}%
  \BibitemOpen
  \bibfield  {author} {\bibinfo {author} {\bibfnamefont {J.}~\bibnamefont {Feather}}, \bibinfo {author} {\bibfnamefont {G.}~\bibnamefont {Leclerc}}, \bibinfo {author} {\bibfnamefont {A.}~\bibnamefont {Madry}},\ and\ \bibinfo {author} {\bibfnamefont {J.~H.}\ \bibnamefont {McDermott}},\ }\bibfield  {title} {\enquote {\bibinfo {title} {{Model metamers reveal divergent invariances between biological and artificial neural networks}},}\ }\href@noop {} {\bibfield  {journal} {\bibinfo  {journal} {Nat.~Neurosci.}\ }\textbf {\bibinfo {volume} {26}},\ \bibinfo {pages} {2017--2034} (\bibinfo {year} {2023})}\BibitemShut {NoStop}%
\bibitem [{\citenamefont {Cheng}\ \emph {et~al.}(2023)\citenamefont {Cheng}, \citenamefont {Horikawa}, \citenamefont {Majima}, \citenamefont {Tanaka}, \citenamefont {Abdelhack}, \citenamefont {Aoki}, \citenamefont {Hirano},\ and\ \citenamefont {Kamitani}}]{Che23}%
  \BibitemOpen
  \bibfield  {author} {\bibinfo {author} {\bibfnamefont {F.~L.}\ \bibnamefont {Cheng}}, \bibinfo {author} {\bibfnamefont {T.}~\bibnamefont {Horikawa}}, \bibinfo {author} {\bibfnamefont {K.}~\bibnamefont {Majima}}, \bibinfo {author} {\bibfnamefont {M.}~\bibnamefont {Tanaka}}, \bibinfo {author} {\bibfnamefont {M.}~\bibnamefont {Abdelhack}}, \bibinfo {author} {\bibfnamefont {S.~C.}\ \bibnamefont {Aoki}}, \bibinfo {author} {\bibfnamefont {J.}~\bibnamefont {Hirano}},\ and\ \bibinfo {author} {\bibfnamefont {Y.}~\bibnamefont {Kamitani}},\ }\bibfield  {title} {\enquote {\bibinfo {title} {Reconstructing visual illusory experiences from human brain activity},}\ }\href@noop {} {\bibfield  {journal} {\bibinfo  {journal} {Sci.~Adv.}\ }\textbf {\bibinfo {volume} {9}},\ \bibinfo {pages} {eadj3906} (\bibinfo {year} {2023})}\BibitemShut {NoStop}%
\bibitem [{\citenamefont {Kornmeier}\ and\ \citenamefont {Bach}(2005)}]{Kor05}%
  \BibitemOpen
  \bibfield  {author} {\bibinfo {author} {\bibfnamefont {J.}~\bibnamefont {Kornmeier}}\ and\ \bibinfo {author} {\bibfnamefont {M.}~\bibnamefont {Bach}},\ }\bibfield  {title} {\enquote {\bibinfo {title} {The {Necker} cube--an ambiguous figure disambiguated in early visual processing},}\ }\href@noop {} {\bibfield  {journal} {\bibinfo  {journal} {Vision Res.}\ }\textbf {\bibinfo {volume} {45}},\ \bibinfo {pages} {955--960} (\bibinfo {year} {2005})}\BibitemShut {NoStop}%
\bibitem [{\citenamefont {Busemeyer}\ and\ \citenamefont {Bruza}(2012)}]{Bus12}%
  \BibitemOpen
  \bibfield  {author} {\bibinfo {author} {\bibfnamefont {J.~R.}\ \bibnamefont {Busemeyer}}\ and\ \bibinfo {author} {\bibfnamefont {P.~D.}\ \bibnamefont {Bruza}},\ }\href@noop {} {\emph {\bibinfo {title} {Quantum Models of Cognition and Decision}}}\ (\bibinfo  {publisher} {Oxford University Press, New York},\ \bibinfo {year} {2012})\BibitemShut {NoStop}%
\bibitem [{\citenamefont {Pinna}\ \emph {et~al.}(2018)\citenamefont {Pinna}, \citenamefont {Reeves}, \citenamefont {Koenderink}, \citenamefont {{van Doorn}},\ and\ \citenamefont {Deiana}}]{Pin18}%
  \BibitemOpen
  \bibfield  {author} {\bibinfo {author} {\bibfnamefont {B.}~\bibnamefont {Pinna}}, \bibinfo {author} {\bibfnamefont {A.}~\bibnamefont {Reeves}}, \bibinfo {author} {\bibfnamefont {J.}~\bibnamefont {Koenderink}}, \bibinfo {author} {\bibfnamefont {A.}~\bibnamefont {{van Doorn}}},\ and\ \bibinfo {author} {\bibfnamefont {K.}~\bibnamefont {Deiana}},\ }\bibfield  {title} {\enquote {\bibinfo {title} {A new principle of figure-ground segregation: The accentuation},}\ }\href@noop {} {\bibfield  {journal} {\bibinfo  {journal} {Vis.~Res.}\ }\textbf {\bibinfo {volume} {143}},\ \bibinfo {pages} {9--25} (\bibinfo {year} {2018})}\BibitemShut {NoStop}%
\bibitem [{\citenamefont {Khalil}(2021)}]{Kha21_1}%
  \BibitemOpen
  \bibfield  {author} {\bibinfo {author} {\bibfnamefont {E.~L.}\ \bibnamefont {Khalil}},\ }\bibfield  {title} {\enquote {\bibinfo {title} {Why does {Rubin's} vase differ radically from optical illusions? {Framing} effects contra cognitive illusions},}\ }\href {https://doi.org/10.3389/fpsyg.2021.597758} {\bibfield  {journal} {\bibinfo  {journal} {Front.~Psychol.}\ }\textbf {\bibinfo {volume} {12}} (\bibinfo {year} {2021}),\ 10.3389/fpsyg.2021.597758}\BibitemShut {NoStop}%
\bibitem [{\citenamefont {Maksymov}(2024)}]{Mak24_illusions}%
  \BibitemOpen
  \bibfield  {author} {\bibinfo {author} {\bibfnamefont {I.~S.}\ \bibnamefont {Maksymov}},\ }\bibfield  {title} {\enquote {\bibinfo {title} {Quantum-inspired neural network model of optical illusions},}\ }\href@noop {} {\bibfield  {journal} {\bibinfo  {journal} {Algorithms}\ }\textbf {\bibinfo {volume} {17}},\ \bibinfo {pages} {30} (\bibinfo {year} {2024})}\BibitemShut {NoStop}%
\bibitem [{\citenamefont {Inoue}\ and\ \citenamefont {Nakamoto}(1994)}]{Ino94}%
  \BibitemOpen
  \bibfield  {author} {\bibinfo {author} {\bibfnamefont {M.}~\bibnamefont {Inoue}}\ and\ \bibinfo {author} {\bibfnamefont {K.}~\bibnamefont {Nakamoto}},\ }\bibfield  {title} {\enquote {\bibinfo {title} {Dynamics of cognitive interpretations of a {Necker} cube in a chaos neural network},}\ }\href@noop {} {\bibfield  {journal} {\bibinfo  {journal} {Prog.~Theor.~Phys.}\ }\textbf {\bibinfo {volume} {92}},\ \bibinfo {pages} {501--508} (\bibinfo {year} {1994})}\BibitemShut {NoStop}%
\bibitem [{\citenamefont {Kudo}\ \emph {et~al.}(1999)\citenamefont {Kudo}, \citenamefont {Yamamura}, \citenamefont {Ohnishi}, \citenamefont {Kobayashi},\ and\ \citenamefont {Sugie}}]{Kud99}%
  \BibitemOpen
  \bibfield  {author} {\bibinfo {author} {\bibfnamefont {H.}~\bibnamefont {Kudo}}, \bibinfo {author} {\bibfnamefont {T.}~\bibnamefont {Yamamura}}, \bibinfo {author} {\bibfnamefont {N.}~\bibnamefont {Ohnishi}}, \bibinfo {author} {\bibfnamefont {S.}~\bibnamefont {Kobayashi}},\ and\ \bibinfo {author} {\bibfnamefont {N.}~\bibnamefont {Sugie}},\ }\bibfield  {title} {\enquote {\bibinfo {title} {A neural network model of dynamically fluctuatin perception of necker cube as well as dot patterns},}\ }\href@noop {} {\bibfield  {journal} {\bibinfo  {journal} {Proceedings of the American Association for Artificial Intelligence}\ }\textbf {\bibinfo {volume} {16}} (\bibinfo {year} {1999})}\BibitemShut {NoStop}%
\bibitem [{\citenamefont {Matsumoto}, \citenamefont {Xu},\ and\ \citenamefont {Takeno}(2018)}]{Mat18}%
  \BibitemOpen
  \bibfield  {author} {\bibinfo {author} {\bibfnamefont {D.}~\bibnamefont {Matsumoto}}, \bibinfo {author} {\bibfnamefont {H.}~\bibnamefont {Xu}},\ and\ \bibinfo {author} {\bibfnamefont {J.}~\bibnamefont {Takeno}},\ }\bibfield  {title} {\enquote {\bibinfo {title} {Simulation of the cognitive process in looking at {Rubin}'s vase},}\ }\href@noop {} {\bibfield  {journal} {\bibinfo  {journal} {Procedia Comput.~Sci.}\ }\textbf {\bibinfo {volume} {123}},\ \bibinfo {pages} {265--270} (\bibinfo {year} {2018})}\BibitemShut {NoStop}%
\bibitem [{\citenamefont {Araki}, \citenamefont {Tsuruoka},\ and\ \citenamefont {Urakawa}(2020)}]{Ara20}%
  \BibitemOpen
  \bibfield  {author} {\bibinfo {author} {\bibfnamefont {O.}~\bibnamefont {Araki}}, \bibinfo {author} {\bibfnamefont {Y.}~\bibnamefont {Tsuruoka}},\ and\ \bibinfo {author} {\bibfnamefont {T.}~\bibnamefont {Urakawa}},\ }\bibfield  {title} {\enquote {\bibinfo {title} {A neural network model for exogenous perceptual alternations of the {Necker} cube},}\ }\href@noop {} {\bibfield  {journal} {\bibinfo  {journal} {Cogn.~Neurodyn.}\ }\textbf {\bibinfo {volume} {14}},\ \bibinfo {pages} {229--237} (\bibinfo {year} {2020})}\BibitemShut {NoStop}%
\bibitem [{\citenamefont {Sun}\ and\ \citenamefont {Dekel}(2021)}]{Sun21_1}%
  \BibitemOpen
  \bibfield  {author} {\bibinfo {author} {\bibfnamefont {E.~D.}\ \bibnamefont {Sun}}\ and\ \bibinfo {author} {\bibfnamefont {R.}~\bibnamefont {Dekel}},\ }\bibfield  {title} {\enquote {\bibinfo {title} {{ImageNet-trained deep neural networks exhibit illusion-like response to the {Scintillating} grid}},}\ }\href {https://doi.org/10.1167/jov.21.11.15} {\bibfield  {journal} {\bibinfo  {journal} {J.~Vis.}\ }\textbf {\bibinfo {volume} {21}},\ \bibinfo {pages} {15} (\bibinfo {year} {2021})}\BibitemShut {NoStop}%
\bibitem [{\citenamefont {Zhang}, \citenamefont {Yoshida},\ and\ \citenamefont {Li}(2024)}]{Zha24}%
  \BibitemOpen
  \bibfield  {author} {\bibinfo {author} {\bibfnamefont {H.}~\bibnamefont {Zhang}}, \bibinfo {author} {\bibfnamefont {S.}~\bibnamefont {Yoshida}},\ and\ \bibinfo {author} {\bibfnamefont {Z.}~\bibnamefont {Li}},\ }\bibfield  {title} {\enquote {\bibinfo {title} {Brain-like illusion produced by {Skye's Oblique Grating} in deep neural networks},}\ }\href {https://doi.org/10.1371/journal.pone.0299083} {\bibfield  {journal} {\bibinfo  {journal} {PLOS ONE}\ }\textbf {\bibinfo {volume} {19}},\ \bibinfo {pages} {e0299083} (\bibinfo {year} {2024})}\BibitemShut {NoStop}%
\bibitem [{\citenamefont {Zhang}\ and\ \citenamefont {Yoshida}(2024)}]{Zha24_1}%
  \BibitemOpen
  \bibfield  {author} {\bibinfo {author} {\bibfnamefont {H.}~\bibnamefont {Zhang}}\ and\ \bibinfo {author} {\bibfnamefont {S.}~\bibnamefont {Yoshida}},\ }\bibfield  {title} {\enquote {\bibinfo {title} {Exploring deep neural networks in simulating human vision through five optical illusions},}\ }\href@noop {} {\bibfield  {journal} {\bibinfo  {journal} {Appl.~Sci.}\ }\textbf {\bibinfo {volume} {14}},\ \bibinfo {pages} {3429} (\bibinfo {year} {2024})}\BibitemShut {NoStop}%
\bibitem [{\citenamefont {Long}\ and\ \citenamefont {Toppino}(2004)}]{Lon04}%
  \BibitemOpen
  \bibfield  {author} {\bibinfo {author} {\bibfnamefont {G.~M.}\ \bibnamefont {Long}}\ and\ \bibinfo {author} {\bibfnamefont {T.~C.}\ \bibnamefont {Toppino}},\ }\bibfield  {title} {\enquote {\bibinfo {title} {Enduring interest in perceptual ambiguity: alternating views of reversible figures},}\ }\href@noop {} {\bibfield  {journal} {\bibinfo  {journal} {Psychol.~Bull.}\ }\textbf {\bibinfo {volume} {130}},\ \bibinfo {pages} {748--768} (\bibinfo {year} {2004})}\BibitemShut {NoStop}%
\bibitem [{\citenamefont {Carbon}(2014)}]{Car14}%
  \BibitemOpen
  \bibfield  {author} {\bibinfo {author} {\bibfnamefont {C.-C.}\ \bibnamefont {Carbon}},\ }\bibfield  {title} {\enquote {\bibinfo {title} {Understanding human perception by human-made illusions},}\ }\href {https://doi.org/10.3389/fnhum.2014.00566} {\bibfield  {journal} {\bibinfo  {journal} {Front.~Hum.~Neurosci.}\ }\textbf {\bibinfo {volume} {8}} (\bibinfo {year} {2014}),\ 10.3389/fnhum.2014.00566}\BibitemShut {NoStop}%
\bibitem [{\citenamefont {Gomez-Villa}\ \emph {et~al.}(2020)\citenamefont {Gomez-Villa}, \citenamefont {Mart{\'i}n}, \citenamefont {Vazquez-Corral}, \citenamefont {Bertalm{\'i}o},\ and\ \citenamefont {Malo}}]{Gom20}%
  \BibitemOpen
  \bibfield  {author} {\bibinfo {author} {\bibfnamefont {A.}~\bibnamefont {Gomez-Villa}}, \bibinfo {author} {\bibfnamefont {A.}~\bibnamefont {Mart{\'i}n}}, \bibinfo {author} {\bibfnamefont {J.}~\bibnamefont {Vazquez-Corral}}, \bibinfo {author} {\bibfnamefont {M.}~\bibnamefont {Bertalm{\'i}o}},\ and\ \bibinfo {author} {\bibfnamefont {J.}~\bibnamefont {Malo}},\ }\bibfield  {title} {\enquote {\bibinfo {title} {Color illusions also deceive {CNNs} for low-level vision tasks: Analysis and implications},}\ }\href@noop {} {\bibfield  {journal} {\bibinfo  {journal} {Vis.~Res.}\ }\textbf {\bibinfo {volume} {176}},\ \bibinfo {pages} {156--174} (\bibinfo {year} {2020})}\BibitemShut {NoStop}%
\bibitem [{\citenamefont {Kim}\ \emph {et~al.}(2021)\citenamefont {Kim}, \citenamefont {Reif}, \citenamefont {Wattenberg}, \citenamefont {Bengio},\ and\ \citenamefont {Mozer}}]{Kim21}%
  \BibitemOpen
  \bibfield  {author} {\bibinfo {author} {\bibfnamefont {B.}~\bibnamefont {Kim}}, \bibinfo {author} {\bibfnamefont {E.}~\bibnamefont {Reif}}, \bibinfo {author} {\bibfnamefont {M.}~\bibnamefont {Wattenberg}}, \bibinfo {author} {\bibfnamefont {S.}~\bibnamefont {Bengio}},\ and\ \bibinfo {author} {\bibfnamefont {M.~C.}\ \bibnamefont {Mozer}},\ }\bibfield  {title} {\enquote {\bibinfo {title} {Neural networks trained on natural scenes exhibit gestalt closure},}\ }\href@noop {} {\bibfield  {journal} {\bibinfo  {journal} {Comput.~Brain Behav.}\ }\textbf {\bibinfo {volume} {4}},\ \bibinfo {pages} {251--263} (\bibinfo {year} {2021})}\BibitemShut {NoStop}%
\bibitem [{\citenamefont {Glorot}, \citenamefont {Bordes},\ and\ \citenamefont {Bengio}(2011)}]{Glo11}%
  \BibitemOpen
  \bibfield  {author} {\bibinfo {author} {\bibfnamefont {X.}~\bibnamefont {Glorot}}, \bibinfo {author} {\bibfnamefont {A.}~\bibnamefont {Bordes}},\ and\ \bibinfo {author} {\bibfnamefont {Y.}~\bibnamefont {Bengio}},\ }\bibfield  {title} {\enquote {\bibinfo {title} {Deep sparse rectifier neural networks},}\ }in\ \href {https://proceedings.mlr.press/v15/glorot11a.html} {\emph {\bibinfo {booktitle} {Proceedings of the Fourteenth International Conference on Artificial Intelligence and Statistics}}},\ \bibinfo {series} {Proceedings of Machine Learning Research}, Vol.~\bibinfo {volume} {15},\ \bibinfo {editor} {edited by\ \bibinfo {editor} {\bibfnamefont {G.}~\bibnamefont {Gordon}}, \bibinfo {editor} {\bibfnamefont {D.}~\bibnamefont {Dunson}},\ and\ \bibinfo {editor} {\bibfnamefont {M.}~\bibnamefont {Dudík}}}\ (\bibinfo {address} {Fort Lauderdale, FL, USA},\ \bibinfo {year} {2011})\ pp.\ \bibinfo {pages} {315--323}\BibitemShut {NoStop}%
\bibitem [{\citenamefont {Bhumbra}(2018)}]{Bhu18}%
  \BibitemOpen
  \bibfield  {author} {\bibinfo {author} {\bibfnamefont {G.~S.}\ \bibnamefont {Bhumbra}},\ }\href {https://arxiv.org/abs/1804.11237} {\enquote {\bibinfo {title} {Deep learning improved by biological activation functions},}\ } (\bibinfo {year} {2018}),\ \Eprint {https://arxiv.org/abs/1804.11237} {arXiv:1804.11237} \BibitemShut {NoStop}%
\bibitem [{\citenamefont {Jones}\ and\ \citenamefont {Kording}(2021)}]{Jon21}%
  \BibitemOpen
  \bibfield  {author} {\bibinfo {author} {\bibfnamefont {I.~S.}\ \bibnamefont {Jones}}\ and\ \bibinfo {author} {\bibfnamefont {K.~P.}\ \bibnamefont {Kording}},\ }\bibfield  {title} {\enquote {\bibinfo {title} {{Might a single neuron solve interesting machine learning problems through successive computations on its dendritic tree?}}}\ }\href@noop {} {\bibfield  {journal} {\bibinfo  {journal} {Neural Comput.}\ }\textbf {\bibinfo {volume} {33}},\ \bibinfo {pages} {1554--1571} (\bibinfo {year} {2021})}\BibitemShut {NoStop}%
\bibitem [{\citenamefont {Watanabe}\ \emph {et~al.}(2018)\citenamefont {Watanabe}, \citenamefont {Kitaoka}, \citenamefont {Sakamoto}, \citenamefont {Yasugi},\ and\ \citenamefont {Tanaka}}]{Wat18}%
  \BibitemOpen
  \bibfield  {author} {\bibinfo {author} {\bibfnamefont {E.}~\bibnamefont {Watanabe}}, \bibinfo {author} {\bibfnamefont {A.}~\bibnamefont {Kitaoka}}, \bibinfo {author} {\bibfnamefont {K.}~\bibnamefont {Sakamoto}}, \bibinfo {author} {\bibfnamefont {M.}~\bibnamefont {Yasugi}},\ and\ \bibinfo {author} {\bibfnamefont {K.}~\bibnamefont {Tanaka}},\ }\bibfield  {title} {\enquote {\bibinfo {title} {Illusory motion reproduced by deep neural networks trained for prediction},}\ }\href {https://doi.org/10.3389/fpsyg.2018.00345} {\bibfield  {journal} {\bibinfo  {journal} {Front.~Psychol.}\ }\textbf {\bibinfo {volume} {9}} (\bibinfo {year} {2018}),\ 10.3389/fpsyg.2018.00345}\BibitemShut {NoStop}%
\bibitem [{\citenamefont {M{\'e}ly}, \citenamefont {Linsley},\ and\ \citenamefont {Serre}(2018)}]{Mel18}%
  \BibitemOpen
  \bibfield  {author} {\bibinfo {author} {\bibfnamefont {D.~A.}\ \bibnamefont {M{\'e}ly}}, \bibinfo {author} {\bibfnamefont {D.}~\bibnamefont {Linsley}},\ and\ \bibinfo {author} {\bibfnamefont {T.}~\bibnamefont {Serre}},\ }\bibfield  {title} {\enquote {\bibinfo {title} {Complementary surrounds explain diverse contextual phenomena across visual modalities},}\ }\href@noop {} {\bibfield  {journal} {\bibinfo  {journal} {Psychol.~Rev.}\ }\textbf {\bibinfo {volume} {125}},\ \bibinfo {pages} {769--784} (\bibinfo {year} {2018})}\BibitemShut {NoStop}%
\bibitem [{\citenamefont {Kubota}, \citenamefont {Hiyama},\ and\ \citenamefont {Inami}(2021)}]{Kub21_1}%
  \BibitemOpen
  \bibfield  {author} {\bibinfo {author} {\bibfnamefont {Y.}~\bibnamefont {Kubota}}, \bibinfo {author} {\bibfnamefont {A.}~\bibnamefont {Hiyama}},\ and\ \bibinfo {author} {\bibfnamefont {M.}~\bibnamefont {Inami}},\ }\bibfield  {title} {\enquote {\bibinfo {title} {A machine learning model perceiving brightness optical illusions: Quantitative evaluation with psychophysical data},}\ }in\ \href {https://doi.org/10.1145/3458709.3458952} {\emph {\bibinfo {booktitle} {Proceedings of the Augmented Humans International Conference 2021}}}\ (\bibinfo {year} {2021})\ pp.\ \bibinfo {pages} {174--182}\BibitemShut {NoStop}%
\bibitem [{\citenamefont {Shahgir}\ \emph {et~al.}(2024)\citenamefont {Shahgir}, \citenamefont {Sayeed}, \citenamefont {Bhattacharjee}, \citenamefont {Ahmad}, \citenamefont {Dong},\ and\ \citenamefont {Shahriyar}}]{Sha24}%
  \BibitemOpen
  \bibfield  {author} {\bibinfo {author} {\bibfnamefont {H.~S.}\ \bibnamefont {Shahgir}}, \bibinfo {author} {\bibfnamefont {K.~S.}\ \bibnamefont {Sayeed}}, \bibinfo {author} {\bibfnamefont {A.}~\bibnamefont {Bhattacharjee}}, \bibinfo {author} {\bibfnamefont {W.~U.}\ \bibnamefont {Ahmad}}, \bibinfo {author} {\bibfnamefont {Y.}~\bibnamefont {Dong}},\ and\ \bibinfo {author} {\bibfnamefont {R.}~\bibnamefont {Shahriyar}},\ }\href@noop {} {\enquote {\bibinfo {title} {{IllusionVQA}: A challenging optical illusion dataset for vision language models},}\ } (\bibinfo {year} {2024}),\ \Eprint {https://arxiv.org/abs/2403.15952} {arXiv:2403.15952} \BibitemShut {NoStop}%
\bibitem [{\citenamefont {Paxton}\ and\ \citenamefont {Smith}(2018)}]{Pax18}%
  \BibitemOpen
  \bibfield  {author} {\bibinfo {author} {\bibfnamefont {A.~B.}\ \bibnamefont {Paxton}}\ and\ \bibinfo {author} {\bibfnamefont {D.}~\bibnamefont {Smith}},\ }\bibfield  {title} {\enquote {\bibinfo {title} {Visual cues from an underwater illusion increase relative abundance of highly reef-associated fish on an artificial reef},}\ }\href@noop {} {\bibfield  {journal} {\bibinfo  {journal} {Mar.~Freshwater Res.}\ }\textbf {\bibinfo {volume} {69}},\ \bibinfo {pages} {614--619} (\bibinfo {year} {2018})}\BibitemShut {NoStop}%
\bibitem [{\citenamefont {Agrillo}\ \emph {et~al.}(2020)\citenamefont {Agrillo}, \citenamefont {Santac{\`a}}, \citenamefont {Pecunioso},\ and\ \citenamefont {Petrazzini}}]{Agr20}%
  \BibitemOpen
  \bibfield  {author} {\bibinfo {author} {\bibfnamefont {C.}~\bibnamefont {Agrillo}}, \bibinfo {author} {\bibfnamefont {M.}~\bibnamefont {Santac{\`a}}}, \bibinfo {author} {\bibfnamefont {A.}~\bibnamefont {Pecunioso}},\ and\ \bibinfo {author} {\bibfnamefont {M.~E.~M.}\ \bibnamefont {Petrazzini}},\ }\bibfield  {title} {\enquote {\bibinfo {title} {Everything is subjective under water surface, too: visual illusions in fish},}\ }\href@noop {} {\bibfield  {journal} {\bibinfo  {journal} {Anim.~Cogn.}\ }\textbf {\bibinfo {volume} {23}},\ \bibinfo {pages} {251--264} (\bibinfo {year} {2020})}\BibitemShut {NoStop}%
\bibitem [{\citenamefont {Gaetz}\ \emph {et~al.}(1998)\citenamefont {Gaetz}, \citenamefont {Weinberg}, \citenamefont {Rzempoluck},\ and\ \citenamefont {Jantzen}}]{Gae98}%
  \BibitemOpen
  \bibfield  {author} {\bibinfo {author} {\bibfnamefont {M.}~\bibnamefont {Gaetz}}, \bibinfo {author} {\bibfnamefont {H.}~\bibnamefont {Weinberg}}, \bibinfo {author} {\bibfnamefont {E.}~\bibnamefont {Rzempoluck}},\ and\ \bibinfo {author} {\bibfnamefont {K.~J.}\ \bibnamefont {Jantzen}},\ }\bibfield  {title} {\enquote {\bibinfo {title} {Neural network classifications and correlation analysis of {EEG} and {MEG} activity accompanying spontaneous reversals of the {Necker} cube},}\ }\href@noop {} {\bibfield  {journal} {\bibinfo  {journal} {Cogn.~Brain Res.}\ }\textbf {\bibinfo {volume} {6}},\ \bibinfo {pages} {335--346} (\bibinfo {year} {1998})}\BibitemShut {NoStop}%
\bibitem [{\citenamefont {Shimaoka}\ \emph {et~al.}(2010)\citenamefont {Shimaoka}, \citenamefont {Kitajo}, \citenamefont {Kaneko},\ and\ \citenamefont {Yamaguchi}}]{Shi10}%
  \BibitemOpen
  \bibfield  {author} {\bibinfo {author} {\bibfnamefont {D.}~\bibnamefont {Shimaoka}}, \bibinfo {author} {\bibfnamefont {K.}~\bibnamefont {Kitajo}}, \bibinfo {author} {\bibfnamefont {K.}~\bibnamefont {Kaneko}},\ and\ \bibinfo {author} {\bibfnamefont {Y.}~\bibnamefont {Yamaguchi}},\ }\bibfield  {title} {\enquote {\bibinfo {title} {Transient process of cortical activity during {Necker} cube perception: from local clusters to global synchrony},}\ }\href@noop {} {\bibfield  {journal} {\bibinfo  {journal} {Nonlinear Biomed.~Phys.}\ }\textbf {\bibinfo {volume} {4}},\ \bibinfo {pages} {S7} (\bibinfo {year} {2010})}\BibitemShut {NoStop}%
\bibitem [{\citenamefont {Piantoni}\ \emph {et~al.}(2017)\citenamefont {Piantoni}, \citenamefont {Romeijn}, \citenamefont {Gomez-Herrero}, \citenamefont {{Van Der Werf}},\ and\ \citenamefont {{Van Someren}}}]{Pia17}%
  \BibitemOpen
  \bibfield  {author} {\bibinfo {author} {\bibfnamefont {G.}~\bibnamefont {Piantoni}}, \bibinfo {author} {\bibfnamefont {N.}~\bibnamefont {Romeijn}}, \bibinfo {author} {\bibfnamefont {G.}~\bibnamefont {Gomez-Herrero}}, \bibinfo {author} {\bibfnamefont {Y.~D.}\ \bibnamefont {{Van Der Werf}}},\ and\ \bibinfo {author} {\bibfnamefont {E.~J.~W.}\ \bibnamefont {{Van Someren}}},\ }\bibfield  {title} {\enquote {\bibinfo {title} {Alpha power predicts persistence of bistable perception},}\ }\href@noop {} {\bibfield  {journal} {\bibinfo  {journal} {Sci.~Rep.}\ }\textbf {\bibinfo {volume} {7}},\ \bibinfo {pages} {5208} (\bibinfo {year} {2017})}\BibitemShut {NoStop}%
\bibitem [{\citenamefont {Joos}\ \emph {et~al.}(2020)\citenamefont {Joos}, \citenamefont {Giersch}, \citenamefont {Hecker}, \citenamefont {Schipp}, \citenamefont {Heinrich}, \citenamefont {{van Elst}},\ and\ \citenamefont {Kornmeier}}]{Joo20}%
  \BibitemOpen
  \bibfield  {author} {\bibinfo {author} {\bibfnamefont {E.}~\bibnamefont {Joos}}, \bibinfo {author} {\bibfnamefont {A.}~\bibnamefont {Giersch}}, \bibinfo {author} {\bibfnamefont {L.}~\bibnamefont {Hecker}}, \bibinfo {author} {\bibfnamefont {J.}~\bibnamefont {Schipp}}, \bibinfo {author} {\bibfnamefont {S.~P.}\ \bibnamefont {Heinrich}}, \bibinfo {author} {\bibfnamefont {L.~T.}\ \bibnamefont {{van Elst}}},\ and\ \bibinfo {author} {\bibfnamefont {J.}~\bibnamefont {Kornmeier}},\ }\bibfield  {title} {\enquote {\bibinfo {title} {Large {EEG} amplitude effects are highly similar across {Necker} cube, smiley, and abstract stimuli},}\ }\href@noop {} {\bibfield  {journal} {\bibinfo  {journal} {PLoS ONE}\ }\textbf {\bibinfo {volume} {15}},\ \bibinfo {pages} {e0232928} (\bibinfo {year} {2020})}\BibitemShut {NoStop}%
\bibitem [{\citenamefont {Matsumiya}\ and\ \citenamefont {Furukawa}(2023)}]{Mat23}%
  \BibitemOpen
  \bibfield  {author} {\bibinfo {author} {\bibfnamefont {K.}~\bibnamefont {Matsumiya}}\ and\ \bibinfo {author} {\bibfnamefont {S.}~\bibnamefont {Furukawa}},\ }\bibfield  {title} {\enquote {\bibinfo {title} {Perceptual decisions interfere more with eye movements than with reach movements},}\ }\href@noop {} {\bibfield  {journal} {\bibinfo  {journal} {Commun.~Biol.}\ }\textbf {\bibinfo {volume} {6}},\ \bibinfo {pages} {882} (\bibinfo {year} {2023})}\BibitemShut {NoStop}%
\bibitem [{\citenamefont {Atmanspacher}, \citenamefont {Filk},\ and\ \citenamefont {R{\"o}mer}(2004)}]{Atm04}%
  \BibitemOpen
  \bibfield  {author} {\bibinfo {author} {\bibfnamefont {H.}~\bibnamefont {Atmanspacher}}, \bibinfo {author} {\bibfnamefont {T.}~\bibnamefont {Filk}},\ and\ \bibinfo {author} {\bibfnamefont {H.}~\bibnamefont {R{\"o}mer}},\ }\bibfield  {title} {\enquote {\bibinfo {title} {Quantum {Zeno} features of bistable perception},}\ }\href@noop {} {\bibfield  {journal} {\bibinfo  {journal} {Biol.~Cybern.}\ }\textbf {\bibinfo {volume} {90}},\ \bibinfo {pages} {33--40} (\bibinfo {year} {2004})}\BibitemShut {NoStop}%
\bibitem [{\citenamefont {Khrennikov}(2006)}]{Khr06}%
  \BibitemOpen
  \bibfield  {author} {\bibinfo {author} {\bibfnamefont {A.}~\bibnamefont {Khrennikov}},\ }\bibfield  {title} {\enquote {\bibinfo {title} {Quantum-like brain: ``interference of minds''},}\ }\href@noop {} {\bibfield  {journal} {\bibinfo  {journal} {Biosystems}\ }\textbf {\bibinfo {volume} {84}},\ \bibinfo {pages} {225--241} (\bibinfo {year} {2006})}\BibitemShut {NoStop}%
\bibitem [{\citenamefont {Pothos}\ and\ \citenamefont {Busemeyer}(2022)}]{Pot22}%
  \BibitemOpen
  \bibfield  {author} {\bibinfo {author} {\bibfnamefont {E.~M.}\ \bibnamefont {Pothos}}\ and\ \bibinfo {author} {\bibfnamefont {J.~R.}\ \bibnamefont {Busemeyer}},\ }\bibfield  {title} {\enquote {\bibinfo {title} {Quantum cognition},}\ }\href@noop {} {\bibfield  {journal} {\bibinfo  {journal} {Annu.~Rev.~Psychol.}\ }\textbf {\bibinfo {volume} {73}},\ \bibinfo {pages} {749--778} (\bibinfo {year} {2022})}\BibitemShut {NoStop}%
\bibitem [{\citenamefont {Liaci}\ \emph {et~al.}(2018)\citenamefont {Liaci}, \citenamefont {Fischer}, \citenamefont {Atmanspacher}, \citenamefont {Heinrichs}, \citenamefont {Tebartz~van Elst},\ and\ \citenamefont {Kornmeier}}]{Lia18}%
  \BibitemOpen
  \bibfield  {author} {\bibinfo {author} {\bibfnamefont {E.}~\bibnamefont {Liaci}}, \bibinfo {author} {\bibfnamefont {A.}~\bibnamefont {Fischer}}, \bibinfo {author} {\bibfnamefont {H.}~\bibnamefont {Atmanspacher}}, \bibinfo {author} {\bibfnamefont {M.}~\bibnamefont {Heinrichs}}, \bibinfo {author} {\bibfnamefont {L.}~\bibnamefont {Tebartz~van Elst}},\ and\ \bibinfo {author} {\bibfnamefont {J.}~\bibnamefont {Kornmeier}},\ }\bibfield  {title} {\enquote {\bibinfo {title} {Positive and negative hysteresis effects for the perception of geometric and emotional ambiguities},}\ }\href {https://doi.org/10.1371/journal.pone.0202398} {\bibfield  {journal} {\bibinfo  {journal} {PLOS ONE}\ }\textbf {\bibinfo {volume} {13}},\ \bibinfo {pages} {e0202398} (\bibinfo {year} {2018})}\BibitemShut {NoStop}%
\bibitem [{\citenamefont {Stanney}\ \emph {et~al.}(2009)\citenamefont {Stanney}, \citenamefont {Schmorrow}, \citenamefont {Johnston}, \citenamefont {Fuchs}, \citenamefont {Jones}, \citenamefont {Hale}, \citenamefont {Ahmad},\ and\ \citenamefont {Young}}]{Sta09}%
  \BibitemOpen
  \bibfield  {author} {\bibinfo {author} {\bibfnamefont {K.~M.}\ \bibnamefont {Stanney}}, \bibinfo {author} {\bibfnamefont {D.~D.}\ \bibnamefont {Schmorrow}}, \bibinfo {author} {\bibfnamefont {M.}~\bibnamefont {Johnston}}, \bibinfo {author} {\bibfnamefont {S.}~\bibnamefont {Fuchs}}, \bibinfo {author} {\bibfnamefont {D.}~\bibnamefont {Jones}}, \bibinfo {author} {\bibfnamefont {K.~S.}\ \bibnamefont {Hale}}, \bibinfo {author} {\bibfnamefont {A.}~\bibnamefont {Ahmad}},\ and\ \bibinfo {author} {\bibfnamefont {P.}~\bibnamefont {Young}},\ }\bibfield  {title} {\enquote {\bibinfo {title} {Augmented cognition: An overview},}\ }\href@noop {} {\bibfield  {journal} {\bibinfo  {journal} {Rev.~Hum.~Factors Ergon.}\ }\textbf {\bibinfo {volume} {5}},\ \bibinfo {pages} {195--224} (\bibinfo {year} {2009})}\BibitemShut {NoStop}%
\bibitem [{\citenamefont {Agarwal}\ and\ \citenamefont {Dagli}(2013)}]{Aga13}%
  \BibitemOpen
  \bibfield  {author} {\bibinfo {author} {\bibfnamefont {S.}~\bibnamefont {Agarwal}}\ and\ \bibinfo {author} {\bibfnamefont {C.~H.}\ \bibnamefont {Dagli}},\ }\bibfield  {title} {\enquote {\bibinfo {title} {Augmented cognition in human–system interaction through coupled action of body sensor network and agent based modeling},}\ }\href@noop {} {\bibfield  {journal} {\bibinfo  {journal} {Procedia Comput.~Sci.}\ }\textbf {\bibinfo {volume} {16}},\ \bibinfo {pages} {20--28} (\bibinfo {year} {2013})}\BibitemShut {NoStop}%
\bibitem [{\citenamefont {Foster}\ and\ \citenamefont {Efthymiou}(2022)}]{KPMG}%
  \BibitemOpen
  \bibfield  {author} {\bibinfo {author} {\bibfnamefont {C.}~\bibnamefont {Foster}}\ and\ \bibinfo {author} {\bibfnamefont {T.}~\bibnamefont {Efthymiou}},\ }\bibfield  {title} {\enquote {\bibinfo {title} {Operational optimisation: Decision-making beyond human capability},}\ }\href@noop {} {\bibfield  {journal} {\bibinfo  {journal} {KPMG Report}\ ,\ \bibinfo {pages} {1--25}} (\bibinfo {year} {2022})}\BibitemShut {NoStop}%
\bibitem [{\citenamefont {Benedek}\ and\ \citenamefont {Caglioti}(2019)}]{Ben18}%
  \BibitemOpen
  \bibfield  {author} {\bibinfo {author} {\bibfnamefont {G.}~\bibnamefont {Benedek}}\ and\ \bibinfo {author} {\bibfnamefont {G.}~\bibnamefont {Caglioti}},\ }\bibfield  {title} {\enquote {\bibinfo {title} {Graphics and quantum mechanics--the {Necker} cube as a quantum-like two-level system},}\ }in\ \href@noop {} {\emph {\bibinfo {booktitle} {Proceedings of the 18th International Conference on Geometry and Graphics}}},\ \bibinfo {editor} {edited by\ \bibinfo {editor} {\bibfnamefont {L.}~\bibnamefont {Cocchiarella}}}\ (\bibinfo  {publisher} {Springer International Publishing},\ \bibinfo {year} {2019})\ pp.\ \bibinfo {pages} {161--172}\BibitemShut {NoStop}%
\bibitem [{\citenamefont {McQuarrie}\ and\ \citenamefont {Simon}(1997)}]{McQ97}%
  \BibitemOpen
  \bibfield  {author} {\bibinfo {author} {\bibfnamefont {D.~A.}\ \bibnamefont {McQuarrie}}\ and\ \bibinfo {author} {\bibfnamefont {J.~D.}\ \bibnamefont {Simon}},\ }\href@noop {} {\emph {\bibinfo {title} {Physical Chemistry -- A Molecular Approach}}}\ (\bibinfo  {publisher} {Prentice Hall, New York},\ \bibinfo {year} {1997})\BibitemShut {NoStop}%
\bibitem [{\citenamefont {Griffiths}(2004)}]{Gri04}%
  \BibitemOpen
  \bibfield  {author} {\bibinfo {author} {\bibfnamefont {D.~J.}\ \bibnamefont {Griffiths}},\ }\href@noop {} {\emph {\bibinfo {title} {Introduction to Quantum Mechanics}}}\ (\bibinfo  {publisher} {Prentice Hall, New Jersey},\ \bibinfo {year} {2004})\BibitemShut {NoStop}%
\bibitem [{\citenamefont {Abbas}\ and\ \citenamefont {Maksymov}(2024)}]{Abb24}%
  \BibitemOpen
  \bibfield  {author} {\bibinfo {author} {\bibfnamefont {A.~H.}\ \bibnamefont {Abbas}}\ and\ \bibinfo {author} {\bibfnamefont {I.~S.}\ \bibnamefont {Maksymov}},\ }\bibfield  {title} {\enquote {\bibinfo {title} {Reservoir computing using measurement-controlled quantum dynamics},}\ }\href@noop {} {\bibfield  {journal} {\bibinfo  {journal} {Electronics}\ }\textbf {\bibinfo {volume} {13}},\ \bibinfo {pages} {1164} (\bibinfo {year} {2024})}\BibitemShut {NoStop}%
\bibitem [{\citenamefont {Symul}, \citenamefont {Assad},\ and\ \citenamefont {Lam}(2011)}]{Sym11}%
  \BibitemOpen
  \bibfield  {author} {\bibinfo {author} {\bibfnamefont {T.}~\bibnamefont {Symul}}, \bibinfo {author} {\bibfnamefont {S.~M.}\ \bibnamefont {Assad}},\ and\ \bibinfo {author} {\bibfnamefont {P.~K.}\ \bibnamefont {Lam}},\ }\bibfield  {title} {\enquote {\bibinfo {title} {{Real time demonstration of high bitrate quantum random number generation with coherent laser light}},}\ }\href@noop {} {\bibfield  {journal} {\bibinfo  {journal} {Appl.~Phys.~Lett.}\ }\textbf {\bibinfo {volume} {98}},\ \bibinfo {pages} {231103} (\bibinfo {year} {2011})}\BibitemShut {NoStop}%
\bibitem [{\citenamefont {Onen}\ \emph {et~al.}(2022)\citenamefont {Onen}, \citenamefont {Emond}, \citenamefont {Wang}, \citenamefont {Zhang}, \citenamefont {Ross}, \citenamefont {Li}, \citenamefont {Yildiz},\ and\ \citenamefont {del Alamo}}]{One22}%
  \BibitemOpen
  \bibfield  {author} {\bibinfo {author} {\bibfnamefont {M.}~\bibnamefont {Onen}}, \bibinfo {author} {\bibfnamefont {N.}~\bibnamefont {Emond}}, \bibinfo {author} {\bibfnamefont {B.}~\bibnamefont {Wang}}, \bibinfo {author} {\bibfnamefont {D.}~\bibnamefont {Zhang}}, \bibinfo {author} {\bibfnamefont {F.~M.}\ \bibnamefont {Ross}}, \bibinfo {author} {\bibfnamefont {J.}~\bibnamefont {Li}}, \bibinfo {author} {\bibfnamefont {B.}~\bibnamefont {Yildiz}},\ and\ \bibinfo {author} {\bibfnamefont {J.~A.}\ \bibnamefont {del Alamo}},\ }\bibfield  {title} {\enquote {\bibinfo {title} {Nanosecond protonic programmable resistors for analog deep learning},}\ }\href@noop {} {\bibfield  {journal} {\bibinfo  {journal} {Science}\ }\textbf {\bibinfo {volume} {377}},\ \bibinfo {pages} {539--543} (\bibinfo {year} {2022})}\BibitemShut {NoStop}%
\bibitem [{\citenamefont {Ye}\ \emph {et~al.}(2023)\citenamefont {Ye}, \citenamefont {Cao}, \citenamefont {Yang}, \citenamefont {Zhang}, \citenamefont {Fang}, \citenamefont {Gu},\ and\ \citenamefont {Yang}}]{Ye23}%
  \BibitemOpen
  \bibfield  {author} {\bibinfo {author} {\bibfnamefont {N.}~\bibnamefont {Ye}}, \bibinfo {author} {\bibfnamefont {L.}~\bibnamefont {Cao}}, \bibinfo {author} {\bibfnamefont {L.}~\bibnamefont {Yang}}, \bibinfo {author} {\bibfnamefont {Z.}~\bibnamefont {Zhang}}, \bibinfo {author} {\bibfnamefont {Z.}~\bibnamefont {Fang}}, \bibinfo {author} {\bibfnamefont {Q.}~\bibnamefont {Gu}},\ and\ \bibinfo {author} {\bibfnamefont {G.-Z.}\ \bibnamefont {Yang}},\ }\bibfield  {title} {\enquote {\bibinfo {title} {Improving the robustness of analog deep neural networks through a {Bayes}-optimized noise injection approach},}\ }\href@noop {} {\bibfield  {journal} {\bibinfo  {journal} {Commun.~Eng.}\ }\textbf {\bibinfo {volume} {2}},\ \bibinfo {pages} {25} (\bibinfo {year} {2023})}\BibitemShut {NoStop}%
\bibitem [{\citenamefont {Dehais}\ \emph {et~al.}(2022)\citenamefont {Dehais}, \citenamefont {Ladouce}, \citenamefont {Darmet}, \citenamefont {Nong}, \citenamefont {Ferraro}, \citenamefont {Torre~Tresols}, \citenamefont {Velut},\ and\ \citenamefont {Labedan}}]{Deh22}%
  \BibitemOpen
  \bibfield  {author} {\bibinfo {author} {\bibfnamefont {F.}~\bibnamefont {Dehais}}, \bibinfo {author} {\bibfnamefont {S.}~\bibnamefont {Ladouce}}, \bibinfo {author} {\bibfnamefont {L.}~\bibnamefont {Darmet}}, \bibinfo {author} {\bibfnamefont {T.-V.}\ \bibnamefont {Nong}}, \bibinfo {author} {\bibfnamefont {G.}~\bibnamefont {Ferraro}}, \bibinfo {author} {\bibfnamefont {J.}~\bibnamefont {Torre~Tresols}}, \bibinfo {author} {\bibfnamefont {S.}~\bibnamefont {Velut}},\ and\ \bibinfo {author} {\bibfnamefont {P.}~\bibnamefont {Labedan}},\ }\bibfield  {title} {\enquote {\bibinfo {title} {Dual passive reactive brain-computer interface: a novel approach to human-machine symbiosis},}\ }\href {https://doi.org/10.3389/fnrgo.2022.824780} {\bibfield  {journal} {\bibinfo  {journal} {Front.~Neuroergonomics}\ }\textbf {\bibinfo {volume} {3}} (\bibinfo {year} {2022}),\ 10.3389/fnrgo.2022.824780}\BibitemShut {NoStop}%
\bibitem [{\citenamefont {Ruengchaijatuporn}\ \emph {et~al.}(2022)\citenamefont {Ruengchaijatuporn}, \citenamefont {Chatnuntawech}, \citenamefont {Teerapittayanon}, \citenamefont {Sriswasdi}, \citenamefont {Itthipuripat}, \citenamefont {Hemrungrojn}, \citenamefont {Bunyabukkana}, \citenamefont {Petchlorlian}, \citenamefont {Chunamchai}, \citenamefont {Chotibut},\ and\ \citenamefont {Chunharas}}]{Rue22}%
  \BibitemOpen
  \bibfield  {author} {\bibinfo {author} {\bibfnamefont {N.}~\bibnamefont {Ruengchaijatuporn}}, \bibinfo {author} {\bibfnamefont {I.}~\bibnamefont {Chatnuntawech}}, \bibinfo {author} {\bibfnamefont {S.}~\bibnamefont {Teerapittayanon}}, \bibinfo {author} {\bibfnamefont {S.}~\bibnamefont {Sriswasdi}}, \bibinfo {author} {\bibfnamefont {S.}~\bibnamefont {Itthipuripat}}, \bibinfo {author} {\bibfnamefont {S.}~\bibnamefont {Hemrungrojn}}, \bibinfo {author} {\bibfnamefont {P.}~\bibnamefont {Bunyabukkana}}, \bibinfo {author} {\bibfnamefont {A.}~\bibnamefont {Petchlorlian}}, \bibinfo {author} {\bibfnamefont {S.}~\bibnamefont {Chunamchai}}, \bibinfo {author} {\bibfnamefont {T.}~\bibnamefont {Chotibut}},\ and\ \bibinfo {author} {\bibfnamefont {C.}~\bibnamefont {Chunharas}},\ }\bibfield  {title} {\enquote {\bibinfo {title} {An explainable self-attention deep neural network for detecting mild cognitive impairment using multi-input digital drawing tasks},}\ }\href@noop {} {\bibfield  {journal} {\bibinfo  {journal}
  {Alz.~Res.~Therapy}\ }\textbf {\bibinfo {volume} {14}},\ \bibinfo {pages} {111} (\bibinfo {year} {2022})}\BibitemShut {NoStop}%
\bibitem [{\citenamefont {Tacikowski}, \citenamefont {Fust},\ and\ \citenamefont {Ehrsson}(2020)}]{Tac20}%
  \BibitemOpen
  \bibfield  {author} {\bibinfo {author} {\bibfnamefont {P.}~\bibnamefont {Tacikowski}}, \bibinfo {author} {\bibfnamefont {J.}~\bibnamefont {Fust}},\ and\ \bibinfo {author} {\bibfnamefont {H.~H.}\ \bibnamefont {Ehrsson}},\ }\bibfield  {title} {\enquote {\bibinfo {title} {Fluidity of gender identity induced by illusory body-sex change},}\ }\href@noop {} {\bibfield  {journal} {\bibinfo  {journal} {Sci.~Reps.}\ }\textbf {\bibinfo {volume} {10}},\ \bibinfo {pages} {14385} (\bibinfo {year} {2020})}\BibitemShut {NoStop}%
\bibitem [{\citenamefont {Georgiev}\ and\ \citenamefont {Glazebrook}(2018)}]{Geo18}%
  \BibitemOpen
  \bibfield  {author} {\bibinfo {author} {\bibfnamefont {D.~D.}\ \bibnamefont {Georgiev}}\ and\ \bibinfo {author} {\bibfnamefont {J.~F.}\ \bibnamefont {Glazebrook}},\ }\bibfield  {title} {\enquote {\bibinfo {title} {The quantum physics of synaptic communication via the {SNARE} protein complex},}\ }\href@noop {} {\bibfield  {journal} {\bibinfo  {journal} {Prog.~Biophys.~Mol.}\ }\textbf {\bibinfo {volume} {135}},\ \bibinfo {pages} {16--29} (\bibinfo {year} {2018})}\BibitemShut {NoStop}%
\bibitem [{\citenamefont {Korn}\ and\ \citenamefont {Faure}(2003)}]{Kor03}%
  \BibitemOpen
  \bibfield  {author} {\bibinfo {author} {\bibfnamefont {H.}~\bibnamefont {Korn}}\ and\ \bibinfo {author} {\bibfnamefont {P.}~\bibnamefont {Faure}},\ }\bibfield  {title} {\enquote {\bibinfo {title} {Is there chaos in the brain? {II. E}xperimental evidence and related models},}\ }\href@noop {} {\bibfield  {journal} {\bibinfo  {journal} {C.~R.~Biol.}\ }\textbf {\bibinfo {volume} {326}},\ \bibinfo {pages} {787--840} (\bibinfo {year} {2003})}\BibitemShut {NoStop}%
\bibitem [{\citenamefont {Tanaka}\ \emph {et~al.}(2019)\citenamefont {Tanaka}, \citenamefont {Yamane}, \citenamefont {H\'{e}roux}, \citenamefont {Nakane}, \citenamefont {Kanazawa}, \citenamefont {Takeda}, \citenamefont {Numata}, \citenamefont {Nakano},\ and\ \citenamefont {Hirose}}]{Tan19}%
  \BibitemOpen
  \bibfield  {author} {\bibinfo {author} {\bibfnamefont {G.}~\bibnamefont {Tanaka}}, \bibinfo {author} {\bibfnamefont {T.}~\bibnamefont {Yamane}}, \bibinfo {author} {\bibfnamefont {J.~B.}\ \bibnamefont {H\'{e}roux}}, \bibinfo {author} {\bibfnamefont {R.}~\bibnamefont {Nakane}}, \bibinfo {author} {\bibfnamefont {N.}~\bibnamefont {Kanazawa}}, \bibinfo {author} {\bibfnamefont {S.}~\bibnamefont {Takeda}}, \bibinfo {author} {\bibfnamefont {H.}~\bibnamefont {Numata}}, \bibinfo {author} {\bibfnamefont {D.}~\bibnamefont {Nakano}},\ and\ \bibinfo {author} {\bibfnamefont {A.}~\bibnamefont {Hirose}},\ }\bibfield  {title} {\enquote {\bibinfo {title} {Recent advances in physical reservoir computing: {A} review},}\ }\href@noop {} {\bibfield  {journal} {\bibinfo  {journal} {Neural Newt.}\ }\textbf {\bibinfo {volume} {115}},\ \bibinfo {pages} {100--123} (\bibinfo {year} {2019})}\BibitemShut {NoStop}%
\bibitem [{\citenamefont {Marcucci}, \citenamefont {Pierangeli},\ and\ \citenamefont {Conti}(2020)}]{Mar20}%
  \BibitemOpen
  \bibfield  {author} {\bibinfo {author} {\bibfnamefont {G.}~\bibnamefont {Marcucci}}, \bibinfo {author} {\bibfnamefont {D.}~\bibnamefont {Pierangeli}},\ and\ \bibinfo {author} {\bibfnamefont {C.}~\bibnamefont {Conti}},\ }\bibfield  {title} {\enquote {\bibinfo {title} {Theory of neuromorphic computing by waves: machine learning by rogue waves, dispersive shocks, and solitons},}\ }\href@noop {} {\bibfield  {journal} {\bibinfo  {journal} {Phys.~Rev.~Lett.}\ }\textbf {\bibinfo {volume} {125}},\ \bibinfo {pages} {093901} (\bibinfo {year} {2020})}\BibitemShut {NoStop}%
\bibitem [{\citenamefont {Maksymov}(2023)}]{Mak23_review}%
  \BibitemOpen
  \bibfield  {author} {\bibinfo {author} {\bibfnamefont {I.~S.}\ \bibnamefont {Maksymov}},\ }\bibfield  {title} {\enquote {\bibinfo {title} {Analogue and physical reservoir computing using water waves: Applications in power engineering and beyond},}\ }\href@noop {} {\bibfield  {journal} {\bibinfo  {journal} {Energies}\ }\textbf {\bibinfo {volume} {16}},\ \bibinfo {pages} {5366} (\bibinfo {year} {2023})}\BibitemShut {NoStop}%
\bibitem [{\citenamefont {Wang}\ \emph {et~al.}(2020)\citenamefont {Wang}, \citenamefont {Wang}, \citenamefont {Wang}, \citenamefont {Wang}, \citenamefont {Li}, \citenamefont {Pan}, \citenamefont {Dai}, \citenamefont {Gao}, \citenamefont {Liu}, \citenamefont {Liu}, \citenamefont {Yang}, \citenamefont {Liu}, \citenamefont {Cheng}, \citenamefont {Chen}, \citenamefont {Wang}, \citenamefont {Watanabe}, \citenamefont {Taniguchi}, \citenamefont {Liang},\ and\ \citenamefont {Miao}}]{Wan20}%
  \BibitemOpen
  \bibfield  {author} {\bibinfo {author} {\bibfnamefont {S.}~\bibnamefont {Wang}}, \bibinfo {author} {\bibfnamefont {C.-Y.}\ \bibnamefont {Wang}}, \bibinfo {author} {\bibfnamefont {P.}~\bibnamefont {Wang}}, \bibinfo {author} {\bibfnamefont {C.}~\bibnamefont {Wang}}, \bibinfo {author} {\bibfnamefont {Z.-A.}\ \bibnamefont {Li}}, \bibinfo {author} {\bibfnamefont {C.}~\bibnamefont {Pan}}, \bibinfo {author} {\bibfnamefont {Y.}~\bibnamefont {Dai}}, \bibinfo {author} {\bibfnamefont {A.}~\bibnamefont {Gao}}, \bibinfo {author} {\bibfnamefont {C.}~\bibnamefont {Liu}}, \bibinfo {author} {\bibfnamefont {J.}~\bibnamefont {Liu}}, \bibinfo {author} {\bibfnamefont {H.}~\bibnamefont {Yang}}, \bibinfo {author} {\bibfnamefont {X.}~\bibnamefont {Liu}}, \bibinfo {author} {\bibfnamefont {B.}~\bibnamefont {Cheng}}, \bibinfo {author} {\bibfnamefont {K.}~\bibnamefont {Chen}}, \bibinfo {author} {\bibfnamefont {Z.}~\bibnamefont {Wang}}, \bibinfo {author} {\bibfnamefont {K.}~\bibnamefont {Watanabe}}, \bibinfo {author} {\bibfnamefont
  {T.}~\bibnamefont {Taniguchi}}, \bibinfo {author} {\bibfnamefont {S.-J.}\ \bibnamefont {Liang}},\ and\ \bibinfo {author} {\bibfnamefont {F.}~\bibnamefont {Miao}},\ }\bibfield  {title} {\enquote {\bibinfo {title} {{Networking retinomorphic sensor with memristive crossbar for brain-inspired visual perception}},}\ }\href@noop {} {\bibfield  {journal} {\bibinfo  {journal} {Natl.~Sci.~Rev}\ }\textbf {\bibinfo {volume} {8}},\ \bibinfo {pages} {nwaa172} (\bibinfo {year} {2020})}\BibitemShut {NoStop}%
\bibitem [{\citenamefont {Yang}, \citenamefont {Intoy},\ and\ \citenamefont {Rucci}(2024)}]{Yan24}%
  \BibitemOpen
  \bibfield  {author} {\bibinfo {author} {\bibfnamefont {B.}~\bibnamefont {Yang}}, \bibinfo {author} {\bibfnamefont {J.}~\bibnamefont {Intoy}},\ and\ \bibinfo {author} {\bibfnamefont {M.}~\bibnamefont {Rucci}},\ }\bibfield  {title} {\enquote {\bibinfo {title} {Eye blinks as a visual processing stage},}\ }\href@noop {} {\bibfield  {journal} {\bibinfo  {journal} {PNAS}\ }\textbf {\bibinfo {volume} {121}},\ \bibinfo {pages} {e2310291121} (\bibinfo {year} {2024})}\BibitemShut {NoStop}%
\bibitem [{\citenamefont {Sakai}\ \emph {et~al.}(1995)\citenamefont {Sakai}, \citenamefont {Katayama}, \citenamefont {Wada},\ and\ \citenamefont {Oiwa}}]{Sak95}%
  \BibitemOpen
  \bibfield  {author} {\bibinfo {author} {\bibfnamefont {K.}~\bibnamefont {Sakai}}, \bibinfo {author} {\bibfnamefont {T.}~\bibnamefont {Katayama}}, \bibinfo {author} {\bibfnamefont {S.}~\bibnamefont {Wada}},\ and\ \bibinfo {author} {\bibfnamefont {K.}~\bibnamefont {Oiwa}},\ }\bibfield  {title} {\enquote {\bibinfo {title} {Chaos causes perspective reversals for ambiguious patterns},}\ }in\ \href@noop {} {\emph {\bibinfo {booktitle} {Advances in Intelligent Computing --- IPMU '94}}},\ \bibinfo {editor} {edited by\ \bibinfo {editor} {\bibfnamefont {B.}~\bibnamefont {Bouchon-Meunier}}, \bibinfo {editor} {\bibfnamefont {R.~R.}\ \bibnamefont {Yager}},\ and\ \bibinfo {editor} {\bibfnamefont {L.~A.}\ \bibnamefont {Zadeh}}}\ (\bibinfo  {publisher} {Springer Berlin Heidelberg},\ \bibinfo {address} {Berlin, Heidelberg},\ \bibinfo {year} {1995})\ pp.\ \bibinfo {pages} {463--472}\BibitemShut {NoStop}%
\bibitem [{\citenamefont {Ang}\ and\ \citenamefont {Maus}(2020)}]{Ang20}%
  \BibitemOpen
  \bibfield  {author} {\bibinfo {author} {\bibfnamefont {J.~W.~A.}\ \bibnamefont {Ang}}\ and\ \bibinfo {author} {\bibfnamefont {G.~W.}\ \bibnamefont {Maus}},\ }\bibfield  {title} {\enquote {\bibinfo {title} {Boosted visual performance after eye blinks},}\ }\href {https://doi.org/10.1167/jov.20.10.2} {\bibfield  {journal} {\bibinfo  {journal} {J.~Vis.}\ }\textbf {\bibinfo {volume} {20}} (\bibinfo {year} {2020}),\ 10.1167/jov.20.10.2}\BibitemShut {NoStop}%
\bibitem [{\citenamefont {Hershman}\ \emph {et~al.}(2024)\citenamefont {Hershman}, \citenamefont {Share}, \citenamefont {Weiss}, \citenamefont {Henik},\ and\ \citenamefont {Shechter}}]{Her24}%
  \BibitemOpen
  \bibfield  {author} {\bibinfo {author} {\bibfnamefont {R.}~\bibnamefont {Hershman}}, \bibinfo {author} {\bibfnamefont {D.~L.}\ \bibnamefont {Share}}, \bibinfo {author} {\bibfnamefont {E.~M.}\ \bibnamefont {Weiss}}, \bibinfo {author} {\bibfnamefont {A.}~\bibnamefont {Henik}},\ and\ \bibinfo {author} {\bibfnamefont {A.}~\bibnamefont {Shechter}},\ }\bibfield  {title} {\enquote {\bibinfo {title} {Insights from eye blinks into the cognitive processes involved in visual word recognition},}\ }\href {https://doi.org/10.5334/joc.343} {\bibfield  {journal} {\bibinfo  {journal} {J.~Cogn.}\ }\textbf {\bibinfo {volume} {7}},\ \bibinfo {pages} {14} (\bibinfo {year} {2024})}\BibitemShut {NoStop}%
\bibitem [{\citenamefont {Harezlak}\ and\ \citenamefont {Kasprowski}(2018)}]{Har18}%
  \BibitemOpen
  \bibfield  {author} {\bibinfo {author} {\bibfnamefont {K.}~\bibnamefont {Harezlak}}\ and\ \bibinfo {author} {\bibfnamefont {P.}~\bibnamefont {Kasprowski}},\ }\bibfield  {title} {\enquote {\bibinfo {title} {Searching for chaos evidence in eye movement signals},}\ }\href {https://doi.org/10.3390/e20010032} {\bibfield  {journal} {\bibinfo  {journal} {Entropy}\ }\textbf {\bibinfo {volume} {20}},\ \bibinfo {pages} {32} (\bibinfo {year} {2018})}\BibitemShut {NoStop}%
\bibitem [{\citenamefont {Gladilin}\ and\ \citenamefont {Eils}(2015)}]{Gla15}%
  \BibitemOpen
  \bibfield  {author} {\bibinfo {author} {\bibfnamefont {E.}~\bibnamefont {Gladilin}}\ and\ \bibinfo {author} {\bibfnamefont {R.}~\bibnamefont {Eils}},\ }\bibfield  {title} {\enquote {\bibinfo {title} {On the role of spatial phase and phase correlation in vision, illusion, and cognition},}\ }\href {https://doi.org/10.3389/fncom.2015.00045} {\bibfield  {journal} {\bibinfo  {journal} {Front.~Comput.~Neurosci.}\ }\textbf {\bibinfo {volume} {9}} (\bibinfo {year} {2015}),\ 10.3389/fncom.2015.00045}\BibitemShut {NoStop}%
\bibitem [{\citenamefont {Parzefall}\ \emph {et~al.}(2019)\citenamefont {Parzefall}, \citenamefont {Szab{\'o}}, \citenamefont {Taniguchi}, \citenamefont {Watanabe}, \citenamefont {Luisier},\ and\ \citenamefont {Novotny}}]{Par19}%
  \BibitemOpen
  \bibfield  {author} {\bibinfo {author} {\bibfnamefont {M.}~\bibnamefont {Parzefall}}, \bibinfo {author} {\bibfnamefont {{\'A}.}~\bibnamefont {Szab{\'o}}}, \bibinfo {author} {\bibfnamefont {T.}~\bibnamefont {Taniguchi}}, \bibinfo {author} {\bibfnamefont {K.}~\bibnamefont {Watanabe}}, \bibinfo {author} {\bibfnamefont {M.}~\bibnamefont {Luisier}},\ and\ \bibinfo {author} {\bibfnamefont {L.}~\bibnamefont {Novotny}},\ }\bibfield  {title} {\enquote {\bibinfo {title} {Light from van der {Waals} quantum tunneling devices},}\ }\href@noop {} {\bibfield  {journal} {\bibinfo  {journal} {Nat.~Commun.}\ }\textbf {\bibinfo {volume} {10}},\ \bibinfo {pages} {292} (\bibinfo {year} {2019})}\BibitemShut {NoStop}%
\bibitem [{\citenamefont {Fan}\ and\ \citenamefont {Wang}(2018)}]{Fan18}%
  \BibitemOpen
  \bibfield  {author} {\bibinfo {author} {\bibfnamefont {F.}~\bibnamefont {Fan}}\ and\ \bibinfo {author} {\bibfnamefont {G.}~\bibnamefont {Wang}},\ }\bibfield  {title} {\enquote {\bibinfo {title} {Learning from pseudo-randomness with an artificial neural network---{Does God} play pseudo-dice?}}\ }\href@noop {} {\bibfield  {journal} {\bibinfo  {journal} {IEEE Access}\ }\textbf {\bibinfo {volume} {6}},\ \bibinfo {pages} {22987--22992} (\bibinfo {year} {2018})}\BibitemShut {NoStop}%
\bibitem [{\citenamefont {Kornmeier}\ and\ \citenamefont {Bach}(2012)}]{Kor12}%
  \BibitemOpen
  \bibfield  {author} {\bibinfo {author} {\bibfnamefont {J.}~\bibnamefont {Kornmeier}}\ and\ \bibinfo {author} {\bibfnamefont {M.}~\bibnamefont {Bach}},\ }\bibfield  {title} {\enquote {\bibinfo {title} {Ambiguous figures – what happens in the brain when perception changes but not the stimulus},}\ }\href {https://doi.org/10.3389/fnhum.2012.00051} {\bibfield  {journal} {\bibinfo  {journal} {Front.~Hum.~Neurosci.}\ }\textbf {\bibinfo {volume} {6}} (\bibinfo {year} {2012}),\ 10.3389/fnhum.2012.00051}\BibitemShut {NoStop}%
\bibitem [{\citenamefont {Wang}\ \emph {et~al.}(2017)\citenamefont {Wang}, \citenamefont {Sang}, \citenamefont {Hao}, \citenamefont {Zhang}, \citenamefont {Bi},\ and\ \citenamefont {Qiu}}]{Wan17_1}%
  \BibitemOpen
  \bibfield  {author} {\bibinfo {author} {\bibfnamefont {X.}~\bibnamefont {Wang}}, \bibinfo {author} {\bibfnamefont {N.}~\bibnamefont {Sang}}, \bibinfo {author} {\bibfnamefont {L.}~\bibnamefont {Hao}}, \bibinfo {author} {\bibfnamefont {Y.}~\bibnamefont {Zhang}}, \bibinfo {author} {\bibfnamefont {T.}~\bibnamefont {Bi}},\ and\ \bibinfo {author} {\bibfnamefont {J.}~\bibnamefont {Qiu}},\ }\bibfield  {title} {\enquote {\bibinfo {title} {Category selectivity of human visual cortex in perception of {Rubin} face–vase illusion},}\ }\href {https://doi.org/10.3389/fpsyg.2017.01543} {\bibfield  {journal} {\bibinfo  {journal} {Front.~Psychol.}\ }\textbf {\bibinfo {volume} {8}} (\bibinfo {year} {2017}),\ 10.3389/fpsyg.2017.01543}\BibitemShut {NoStop}%
\bibitem [{\citenamefont {Ngo}\ \emph {et~al.}(2008)\citenamefont {Ngo}, \citenamefont {Liu}, \citenamefont {Tilley}, \citenamefont {Pettigrew},\ and\ \citenamefont {Miller}}]{Ngo08}%
  \BibitemOpen
  \bibfield  {author} {\bibinfo {author} {\bibfnamefont {T.~T.}\ \bibnamefont {Ngo}}, \bibinfo {author} {\bibfnamefont {G.~B.}\ \bibnamefont {Liu}}, \bibinfo {author} {\bibfnamefont {A.~J.}\ \bibnamefont {Tilley}}, \bibinfo {author} {\bibfnamefont {J.~D.}\ \bibnamefont {Pettigrew}},\ and\ \bibinfo {author} {\bibfnamefont {S.~M.}\ \bibnamefont {Miller}},\ }\bibfield  {title} {\enquote {\bibinfo {title} {The changing face of perceptual rivalry},}\ }\href@noop {} {\bibfield  {journal} {\bibinfo  {journal} {Brain Res.~Bull.}\ }\textbf {\bibinfo {volume} {75}},\ \bibinfo {pages} {610--618} (\bibinfo {year} {2008})}\BibitemShut {NoStop}%
\bibitem [{\citenamefont {Dayan}\ and\ \citenamefont {Abbott}(2001)}]{Day01}%
  \BibitemOpen
  \bibfield  {author} {\bibinfo {author} {\bibfnamefont {P.}~\bibnamefont {Dayan}}\ and\ \bibinfo {author} {\bibfnamefont {L.~F.}\ \bibnamefont {Abbott}},\ }\href@noop {} {\emph {\bibinfo {title} {Theoretical Neuroscience}}}\ (\bibinfo  {publisher} {MIT Press},\ \bibinfo {year} {2001})\BibitemShut {NoStop}%
\bibitem [{\citenamefont {Schwartz}, \citenamefont {Stapp},\ and\ \citenamefont {Beauregard}(2005)}]{Sch05_1}%
  \BibitemOpen
  \bibfield  {author} {\bibinfo {author} {\bibfnamefont {J.~M.}\ \bibnamefont {Schwartz}}, \bibinfo {author} {\bibfnamefont {H.~P.}\ \bibnamefont {Stapp}},\ and\ \bibinfo {author} {\bibfnamefont {M.}~\bibnamefont {Beauregard}},\ }\bibfield  {title} {\enquote {\bibinfo {title} {Quantum physics in neuroscience and psychology: a neurophysical model of mind–brain interaction},}\ }\href@noop {} {\bibfield  {journal} {\bibinfo  {journal} {Philos.~Trans.~R.~Soc.~Lond.: B Biol.~Sci.}\ }\textbf {\bibinfo {volume} {360}},\ \bibinfo {pages} {1309–1327} (\bibinfo {year} {2005})}\BibitemShut {NoStop}%
\bibitem [{\citenamefont {Busemeyer}, \citenamefont {Fakhari},\ and\ \citenamefont {Kvam}(2017)}]{Bus17}%
  \BibitemOpen
  \bibfield  {author} {\bibinfo {author} {\bibfnamefont {J.~R.}\ \bibnamefont {Busemeyer}}, \bibinfo {author} {\bibfnamefont {P.}~\bibnamefont {Fakhari}},\ and\ \bibinfo {author} {\bibfnamefont {P.}~\bibnamefont {Kvam}},\ }\bibfield  {title} {\enquote {\bibinfo {title} {Neural implementation of operations used in quantum cognition},}\ }\href@noop {} {\bibfield  {journal} {\bibinfo  {journal} {Prog.~Biophys.~Mol.}\ }\textbf {\bibinfo {volume} {130}},\ \bibinfo {pages} {53--60} (\bibinfo {year} {2017})}\BibitemShut {NoStop}%
\bibitem [{\citenamefont {Novicky}\ \emph {et~al.}(2024)\citenamefont {Novicky}, \citenamefont {Parr}, \citenamefont {Friston}, \citenamefont {Mirza},\ and\ \citenamefont {Sajid}}]{Nov24}%
  \BibitemOpen
  \bibfield  {author} {\bibinfo {author} {\bibfnamefont {F.}~\bibnamefont {Novicky}}, \bibinfo {author} {\bibfnamefont {T.}~\bibnamefont {Parr}}, \bibinfo {author} {\bibfnamefont {K.}~\bibnamefont {Friston}}, \bibinfo {author} {\bibfnamefont {M.~B.}\ \bibnamefont {Mirza}},\ and\ \bibinfo {author} {\bibfnamefont {N.}~\bibnamefont {Sajid}},\ }\bibfield  {title} {\enquote {\bibinfo {title} {Bistable perception, precision and neuromodulation},}\ }\href@noop {} {\bibfield  {journal} {\bibinfo  {journal} {Cereb.~Cortex.}\ }\textbf {\bibinfo {volume} {34}},\ \bibinfo {pages} {bhad401} (\bibinfo {year} {2024})}\BibitemShut {NoStop}%
\bibitem [{\citenamefont {Salvador}\ and\ \citenamefont {Chan}(2007)}]{Sal07}%
  \BibitemOpen
  \bibfield  {author} {\bibinfo {author} {\bibfnamefont {S.}~\bibnamefont {Salvador}}\ and\ \bibinfo {author} {\bibfnamefont {P.}~\bibnamefont {Chan}},\ }\bibfield  {title} {\enquote {\bibinfo {title} {Fastdtw: Toward accurate dynamic time warping in linear time and space},}\ }\href@noop {} {\bibfield  {journal} {\bibinfo  {journal} {Intell.~Data Anal.}\ }\textbf {\bibinfo {volume} {11}},\ \bibinfo {pages} {561--580} (\bibinfo {year} {2007})}\BibitemShut {NoStop}%
\bibitem [{\citenamefont {Olsen}, \citenamefont {Markussen},\ and\ \citenamefont {Raket}(2018)}]{Ols18}%
  \BibitemOpen
  \bibfield  {author} {\bibinfo {author} {\bibfnamefont {N.~L.}\ \bibnamefont {Olsen}}, \bibinfo {author} {\bibfnamefont {B.}~\bibnamefont {Markussen}},\ and\ \bibinfo {author} {\bibfnamefont {L.~L.}\ \bibnamefont {Raket}},\ }\bibfield  {title} {\enquote {\bibinfo {title} {{Simultaneous inference for misaligned multivariate functional data}},}\ }\href@noop {} {\bibfield  {journal} {\bibinfo  {journal} {J.~R.~Stat.~Soc., C:~Appl.~Stat.}\ }\textbf {\bibinfo {volume} {67}},\ \bibinfo {pages} {1147--1176} (\bibinfo {year} {2018})}\BibitemShut {NoStop}%
\bibitem [{\citenamefont {Benda}, \citenamefont {Maler},\ and\ \citenamefont {Longtin}(2010)}]{Ben10}%
  \BibitemOpen
  \bibfield  {author} {\bibinfo {author} {\bibfnamefont {J.}~\bibnamefont {Benda}}, \bibinfo {author} {\bibfnamefont {L.}~\bibnamefont {Maler}},\ and\ \bibinfo {author} {\bibfnamefont {A.}~\bibnamefont {Longtin}},\ }\bibfield  {title} {\enquote {\bibinfo {title} {Linear versus nonlinear signal transmission in neuron models with adaptation currents or dynamic thresholds},}\ }\href@noop {} {\bibfield  {journal} {\bibinfo  {journal} {J.~Neurophysiol.}\ }\textbf {\bibinfo {volume} {104}},\ \bibinfo {pages} {2806--2820} (\bibinfo {year} {2010})}\BibitemShut {NoStop}%
\bibitem [{\citenamefont {Lipovetsky}(2018)}]{Lip18}%
  \BibitemOpen
  \bibfield  {author} {\bibinfo {author} {\bibfnamefont {S.}~\bibnamefont {Lipovetsky}},\ }\bibfield  {title} {\enquote {\bibinfo {title} {Quantum paradigm of probability amplitude and complex utility in entangled discrete choice modeling},}\ }\href@noop {} {\bibfield  {journal} {\bibinfo  {journal} {J.~Choice Model.}\ }\textbf {\bibinfo {volume} {27}},\ \bibinfo {pages} {62--73} (\bibinfo {year} {2018})}\BibitemShut {NoStop}%
\bibitem [{\citenamefont {Maksymov}\ and\ \citenamefont {Pogrebna}(2024{\natexlab{b}})}]{Mak24_information1}%
  \BibitemOpen
  \bibfield  {author} {\bibinfo {author} {\bibfnamefont {I.~S.}\ \bibnamefont {Maksymov}}\ and\ \bibinfo {author} {\bibfnamefont {G.}~\bibnamefont {Pogrebna}},\ }\bibfield  {title} {\enquote {\bibinfo {title} {The physics of preference: unravelling imprecision of human preferences through magnetisation dynamics},}\ }\href@noop {} {\bibfield  {journal} {\bibinfo  {journal} {Information}\ }\textbf {\bibinfo {volume} {15}},\ \bibinfo {pages} {413} (\bibinfo {year} {2024}{\natexlab{b}})}\BibitemShut {NoStop}%
\bibitem [{\citenamefont {Zak}(2000)}]{Zak00}%
  \BibitemOpen
  \bibfield  {author} {\bibinfo {author} {\bibfnamefont {M.}~\bibnamefont {Zak}},\ }\bibfield  {title} {\enquote {\bibinfo {title} {Quantum decision-maker},}\ }\href@noop {} {\bibfield  {journal} {\bibinfo  {journal} {Inf.~Sci.}\ }\textbf {\bibinfo {volume} {128}},\ \bibinfo {pages} {199--215} (\bibinfo {year} {2000})}\BibitemShut {NoStop}%
\bibitem [{\citenamefont {Georgiev}(2019)}]{Geo_book}%
  \BibitemOpen
  \bibfield  {author} {\bibinfo {author} {\bibfnamefont {D.~D.}\ \bibnamefont {Georgiev}},\ }\href@noop {} {\emph {\bibinfo {title} {Quantum Information and Consciousness}}}\ (\bibinfo  {publisher} {CRC Press, Boca Raton},\ \bibinfo {year} {2019})\BibitemShut {NoStop}%
\bibitem [{\citenamefont {Georgiev}\ and\ \citenamefont {Glazebrook}(2019)}]{Geo19}%
  \BibitemOpen
  \bibfield  {author} {\bibinfo {author} {\bibfnamefont {D.~D.}\ \bibnamefont {Georgiev}}\ and\ \bibinfo {author} {\bibfnamefont {J.~F.}\ \bibnamefont {Glazebrook}},\ }\bibfield  {title} {\enquote {\bibinfo {title} {Quantum tunneling of {Davydov} solitons through massive barriers},}\ }\href@noop {} {\bibfield  {journal} {\bibinfo  {journal} {Chaos Soliton.~Fract.}\ }\textbf {\bibinfo {volume} {123}},\ \bibinfo {pages} {275--293} (\bibinfo {year} {2019})}\BibitemShut {NoStop}%
\bibitem [{\citenamefont {Georgiev}\ and\ \citenamefont {Glazebrook}(2020)}]{Geo20}%
  \BibitemOpen
  \bibfield  {author} {\bibinfo {author} {\bibfnamefont {D.~D.}\ \bibnamefont {Georgiev}}\ and\ \bibinfo {author} {\bibfnamefont {J.~F.}\ \bibnamefont {Glazebrook}},\ }\bibfield  {title} {\enquote {\bibinfo {title} {Quantum transport and utilization of free energy in protein $\alpha$-helices},}\ }in\ \href@noop {} {\emph {\bibinfo {booktitle} {Quantum Boundaries of Life}}},\ \bibinfo {series} {Advances in Quantum Chemistry}, Vol.~\bibinfo {volume} {82},\ \bibinfo {editor} {edited by\ \bibinfo {editor} {\bibfnamefont {R.~R.}\ \bibnamefont {Pozna{\.n}ski}}\ and\ \bibinfo {editor} {\bibfnamefont {E.~J.}\ \bibnamefont {Br{\"a}ndas}}}\ (\bibinfo  {publisher} {Academic Press},\ \bibinfo {year} {2020})\ pp.\ \bibinfo {pages} {253--300}\BibitemShut {NoStop}%
\bibitem [{\citenamefont {Georgiev}(2021)}]{Geo21}%
  \BibitemOpen
  \bibfield  {author} {\bibinfo {author} {\bibfnamefont {D.~D.}\ \bibnamefont {Georgiev}},\ }\bibfield  {title} {\enquote {\bibinfo {title} {Quantum propensities in the brain cortex and free will},}\ }\href@noop {} {\bibfield  {journal} {\bibinfo  {journal} {Biosystems}\ }\textbf {\bibinfo {volume} {208}},\ \bibinfo {pages} {104474} (\bibinfo {year} {2021})}\BibitemShut {NoStop}%
\bibitem [{\citenamefont {Georgiev}\ and\ \citenamefont {Glazebrook}(2022{\natexlab{a}})}]{Geo22}%
  \BibitemOpen
  \bibfield  {author} {\bibinfo {author} {\bibfnamefont {D.~D.}\ \bibnamefont {Georgiev}}\ and\ \bibinfo {author} {\bibfnamefont {J.~F.}\ \bibnamefont {Glazebrook}},\ }\bibfield  {title} {\enquote {\bibinfo {title} {Thermal stability of solitons in protein $\alpha$-helices},}\ }\href@noop {} {\bibfield  {journal} {\bibinfo  {journal} {Chaos Soliton.~Fract.}\ }\textbf {\bibinfo {volume} {155}},\ \bibinfo {pages} {111644} (\bibinfo {year} {2022}{\natexlab{a}})}\BibitemShut {NoStop}%
\bibitem [{\citenamefont {Georgiev}\ and\ \citenamefont {Glazebrook}(2022{\natexlab{b}})}]{Geo22_1}%
  \BibitemOpen
  \bibfield  {author} {\bibinfo {author} {\bibfnamefont {D.~D.}\ \bibnamefont {Georgiev}}\ and\ \bibinfo {author} {\bibfnamefont {J.~F.}\ \bibnamefont {Glazebrook}},\ }\bibfield  {title} {\enquote {\bibinfo {title} {Quantum tunneling of three-spine solitons through excentric barriers},}\ }\href@noop {} {\bibfield  {journal} {\bibinfo  {journal} {Phys.~Lett.~A}\ }\textbf {\bibinfo {volume} {448}},\ \bibinfo {pages} {128319} (\bibinfo {year} {2022}{\natexlab{b}})}\BibitemShut {NoStop}%
\bibitem [{\citenamefont {Georgiev}\ and\ \citenamefont {Gudder}(2022)}]{Geo22_2}%
  \BibitemOpen
  \bibfield  {author} {\bibinfo {author} {\bibfnamefont {D.~D.}\ \bibnamefont {Georgiev}}\ and\ \bibinfo {author} {\bibfnamefont {S.~P.}\ \bibnamefont {Gudder}},\ }\bibfield  {title} {\enquote {\bibinfo {title} {Sensitivity of entanglement measures in bipartite pure quantum states},}\ }\href@noop {} {\bibfield  {journal} {\bibinfo  {journal} {Mod.~Phys.~Lett.~B}\ }\textbf {\bibinfo {volume} {36}},\ \bibinfo {pages} {2250101} (\bibinfo {year} {2022})}\BibitemShut {NoStop}%
\bibitem [{\citenamefont {Georgiev}(2024)}]{Geo24}%
  \BibitemOpen
  \bibfield  {author} {\bibinfo {author} {\bibfnamefont {D.~D.}\ \bibnamefont {Georgiev}},\ }\bibfield  {title} {\enquote {\bibinfo {title} {Causal potency of consciousness in the physical world},}\ }\href@noop {} {\bibfield  {journal} {\bibinfo  {journal} {Int.~J.~Mod.~Phys.~B}\ }\textbf {\bibinfo {volume} {38}},\ \bibinfo {pages} {2450256} (\bibinfo {year} {2024})}\BibitemShut {NoStop}%
\bibitem [{\citenamefont {Chitambar}\ and\ \citenamefont {Gour}(2019)}]{Chi19}%
  \BibitemOpen
  \bibfield  {author} {\bibinfo {author} {\bibfnamefont {E.}~\bibnamefont {Chitambar}}\ and\ \bibinfo {author} {\bibfnamefont {G.}~\bibnamefont {Gour}},\ }\bibfield  {title} {\enquote {\bibinfo {title} {Quantum resource theories},}\ }\href@noop {} {\bibfield  {journal} {\bibinfo  {journal} {Rev.~Mod.~Phys.}\ }\textbf {\bibinfo {volume} {91}},\ \bibinfo {pages} {025001} (\bibinfo {year} {2019})}\BibitemShut {NoStop}%
\bibitem [{\citenamefont {Khrennikov}\ and\ \citenamefont {Asano}(2020)}]{Khr20}%
  \BibitemOpen
  \bibfield  {author} {\bibinfo {author} {\bibfnamefont {A.}~\bibnamefont {Khrennikov}}\ and\ \bibinfo {author} {\bibfnamefont {M.}~\bibnamefont {Asano}},\ }\bibfield  {title} {\enquote {\bibinfo {title} {A quantum-like model of information processing in the brain},}\ }\href {https://doi.org/10.3390/app10020707} {\bibfield  {journal} {\bibinfo  {journal} {Appl.~Sci.}\ }\textbf {\bibinfo {volume} {10}},\ \bibinfo {pages} {707} (\bibinfo {year} {2020})}\BibitemShut {NoStop}%
\bibitem [{\citenamefont {Markovi{\'c}}\ and\ \citenamefont {Grollier}(2020)}]{Mar20_2}%
  \BibitemOpen
  \bibfield  {author} {\bibinfo {author} {\bibfnamefont {D.}~\bibnamefont {Markovi{\'c}}}\ and\ \bibinfo {author} {\bibfnamefont {J.}~\bibnamefont {Grollier}},\ }\bibfield  {title} {\enquote {\bibinfo {title} {{Quantum neuromorphic computing}},}\ }\href@noop {} {\bibfield  {journal} {\bibinfo  {journal} {Appl.~Phys.~Lett.}\ }\textbf {\bibinfo {volume} {117}},\ \bibinfo {pages} {150501} (\bibinfo {year} {2020})}\BibitemShut {NoStop}%
\end{thebibliography}%

\end{document}